\documentclass[10pt,twocolumn,letterpaper]{article}

\usepackage{iccv}
\usepackage{booktabs}
\usepackage{multirow}
\usepackage{multicol}
\usepackage{times}
\usepackage{epsfig}
\usepackage{graphicx}
\usepackage{amsmath}
\usepackage{amssymb}
\usepackage[inline]{enumitem}
\usepackage{bm}
\usepackage{rotating}
\usepackage{gensymb}
\usepackage{nicefrac}
\usepackage[dvipsnames]{xcolor}
\usepackage{calc}
\usepackage{microtype}  
\usepackage{cuted}  
\usepackage{cite}
\usepackage{float}
\usepackage{caption}
\usepackage{xr}  

\usepackage[pagebackref=true,breaklinks=true,letterpaper=true,colorlinks,bookmarks=false]{hyperref}
\hypersetup{
    citecolor=ForestGreen,
    urlcolor=MidnightBlue,
}

\iccvfinalcopy 
\def\iccvPaperID{10386} 

\newif\ifproceedings  
\proceedingsfalse
\newif\ifsupponly
\supponlyfalse

\ifproceedings
\fi

\ifproceedings
    \newcommand{\supp}{Supplemental} 
\else
    \newcommand{\supp}{Appendix} 
\fi

\DeclareFontFamily{U}{mathb}{}
\DeclareFontShape{U}{mathb}{m}{n}{
  <-5.5> mathb5
  <5.5-6.5> mathb6
  <6.5-7.5> mathb7
  <7.5-8.5> mathb8
  <8.5-9.5> mathb9
  <9.5-11.5> mathb10
  <11.5-> mathb12
}{}
\DeclareSymbolFont{mathb}{U}{mathb}{m}{n}
\DeclareMathSymbol{\drsh}{3}{mathb}{"EB}

\newcommand{\PAR}[1]{\vskip4pt \noindent{\bf #1~}}
\renewcommand{\*}[1]{\bm{\mathrm{#1}}}
\renewcommand{\b}[1]{\textbf{#1}}
\newcommand{\red}[1]{\textcolor{red}{#1}}
\newcommand{\green}[1]{\textcolor{green}{#1}}

\newcommand{\blue}[1]{\textcolor{blue}{#1}}
\newcommand{\0}{\phantom{0}}

\makeatletter
\newcommand{\printfnsymbol}[1]{%
  \textsuperscript{\@fnsymbol{#1}}%
}
\makeatother

\begin{document}

\title{Pixel-Perfect Structure-from-Motion with Featuremetric Refinement}

\author{%
Philipp Lindenberger$^{1}$\thanks{indicates equal contributions}\hspace{.12in}
Paul-Edouard Sarlin$^{2}$\printfnsymbol{1}\hspace{.12in}
Viktor Larsson$^{2}$\hspace{.12in}
Marc Pollefeys$^{2,3}$
\vspace{0.05in}\\
Departments of $^{1}$Mathematics and $^{2}$Computer Science, ETH Zurich\hspace{.2in}
$^{3}$Microsoft
}

\ifsupponly\else 

\maketitle
\ifproceedings\ificcvfinal\thispagestyle{empty}\fi\fi
\ifproceedings\ificcvfinal\pagestyle{empty}\fi\fi

\begin{abstract}
Finding local features that are repeatable across multiple views is a cornerstone of sparse 3D reconstruction.
The classical image matching paradigm detects keypoints per-image once and for all, which can yield poorly-localized features and propagate large errors to the final geometry.
In this paper, we refine two key steps of structure-from-motion by a direct alignment of low-level image information from multiple views: we first adjust the initial keypoint locations prior to any geometric estimation, and subsequently refine points and camera poses as a post-processing.
This refinement is robust to large detection noise and appearance changes, as it optimizes a \emph{featuremetric} error based on dense features predicted by a neural network.
This significantly improves the accuracy of camera poses and scene geometry for a wide range of keypoint detectors, challenging viewing conditions, and off-the-shelf deep features.
Our system easily scales to large image collections, enabling pixel-perfect crowd-sourced localization at scale.
Our code is publicly available at~\href{https://github.com/cvg/pixel-perfect-sfm}{\texttt{github.com/cvg/pixel-perfect-sfm}} as an add-on to the popular SfM software COLMAP.
\end{abstract}

\section{Introduction}

Mapping the world is an important requirement for spatial intelligence applications in augmented reality or robotics.
Tasks like visual localization or path planning can benefit from accurate sparse or dense 3D reconstructions of the environment.
These can be built from images using Structure-from-Motion (SfM), which associates observations across views to estimate camera parameters and 3D scene geometry.
Sparse reconstruction based on matching local image features~\cite{harris1988combined, lowe2004distinctive, bay2006surf, superpoint, dusmanu2019d2, revaud2019r2d2, pautrat2020online, sarlin2020superglue} is the most common due to its scalability and its robustness to appearance changes introduced by varying devices, viewpoints, and temporal conditions found in crowdsourced scenarios~\cite{heinly2015reconstructing, li2012worldwide, irschara2009structure, liu2017efficient, radenovic2016dusk, agarwal2011building, frahm2010building}.

\begin{figure}[t]
    \centering
    \includegraphics[width=0.9\linewidth]{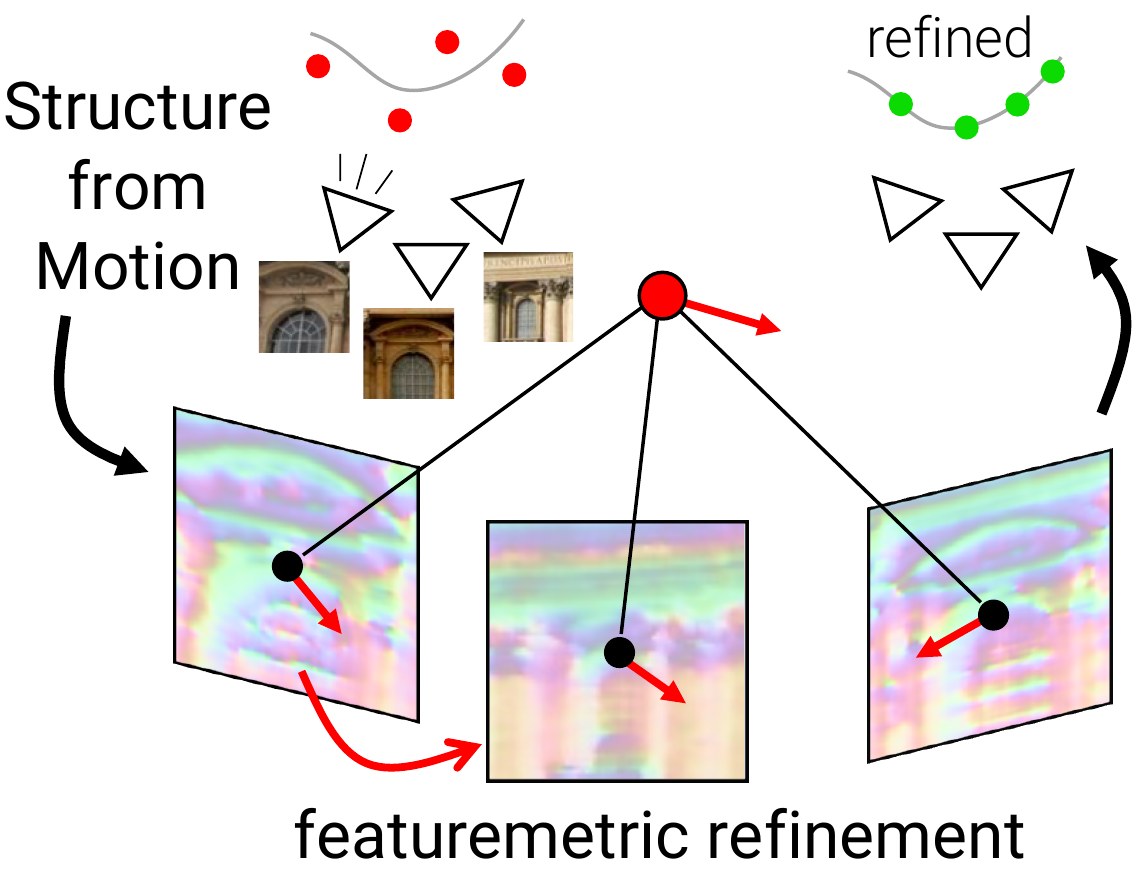}
    \caption{\textbf{From sparse to dense.}
    We improve the accuracy of sparse Structure-from-Motion by refining 2D keypoints, camera poses, and 3D points using the direct alignment of deep features.
    This featuremetric optimization leverages dense image information but can scale to scenes with thousands of images.
    Such refinement results in subpixel-accurate reconstructions, even in challenging conditions.
    }%
    \label{fig:teaser}%
\end{figure}

SfM assumes that sparse interest points~\cite{lowe2004distinctive, harris1988combined, bay2006surf, rosten2006machine, yi2016lift, superpoint, dusmanu2019d2, revaud2019r2d2, tyszkiewicz2020disk} can be reliably detected across views.
It~typically selects such points for each image independently and relies on these initial detections for the remainder of the reconstruction process. 
However, detecting keypoints from a single view is inherently inaccurate due to appearance changes and discrete image sampling~\cite{germain2020s2dnet}.
The advent of convolutional neural network (CNNs) for detection has magnified this issue, as they generally do not retain local image information and instead favor global context.

Multi-view geometric optimization with bundle adjustment~\cite{triggs1999bundle, agarwal2010bundle, jeong2011pushing} is commonly used to refine cameras and points using reprojection errors.
Dusmanu~\etal~\cite{dusmanu2020} proposed to refine keypoint locations prior to SfM via an analogous geometric cost constrained with local optical flow.
This can improve SfM, but has limited accuracy and scalability.

In this work, we argue that local image information is valuable throughout the SfM process to improve its accuracy.
We adjust both keypoints and bundles, before and after reconstruction, by direct image alignment~\cite{LK, engel2017direct, Czarnowski_2017_ICCV} in a learned feature space.
Exploiting this locally-dense information is significantly more accurate than geometric optimization, while deep, high-dimensional features extracted by a CNN ensure wider convergence in challenging conditions.
This formulation elegantly combines globally-discriminative sparse matching with locally-accurate dense details.
It is applicable to both incremental~\cite{snavely2008modeling, schoenberger2016sfm} and global~\cite{martinec2007robust, chatterjee2013efficient, barath2020efficient} SfM irrespective of the types of sparse or dense features.

We validate our approach in experiments evaluating the accuracy of both 3D structure and camera poses in various conditions.
We demonstrate drastic improvements for multiple hand-crafted and learned local features using off-the-shelf CNNs.
The resulting system produces accurate reconstructions and scales well to large scenes with thousands of images.
In the context of visual localization, it can, in addition to providing a more accurate map, also refine poses of single query images with minimal overhead.

For the benefit of the research community, we will release our code as an extension to COLMAP~\cite{schoenberger2016sfm, schoenberger2016mvs} and to the popular localization toolbox hloc~\cite{hloc, sarlin2019coarse}.
We believe that our featuremetric refinement can significantly improve the accuracy of existing datasets~\cite{sattler2018benchmarking} and push the community towards sub-pixel accurate localization at large scale.

\begin{figure}[t]
    \centering
    \includegraphics[width=1.0\linewidth]{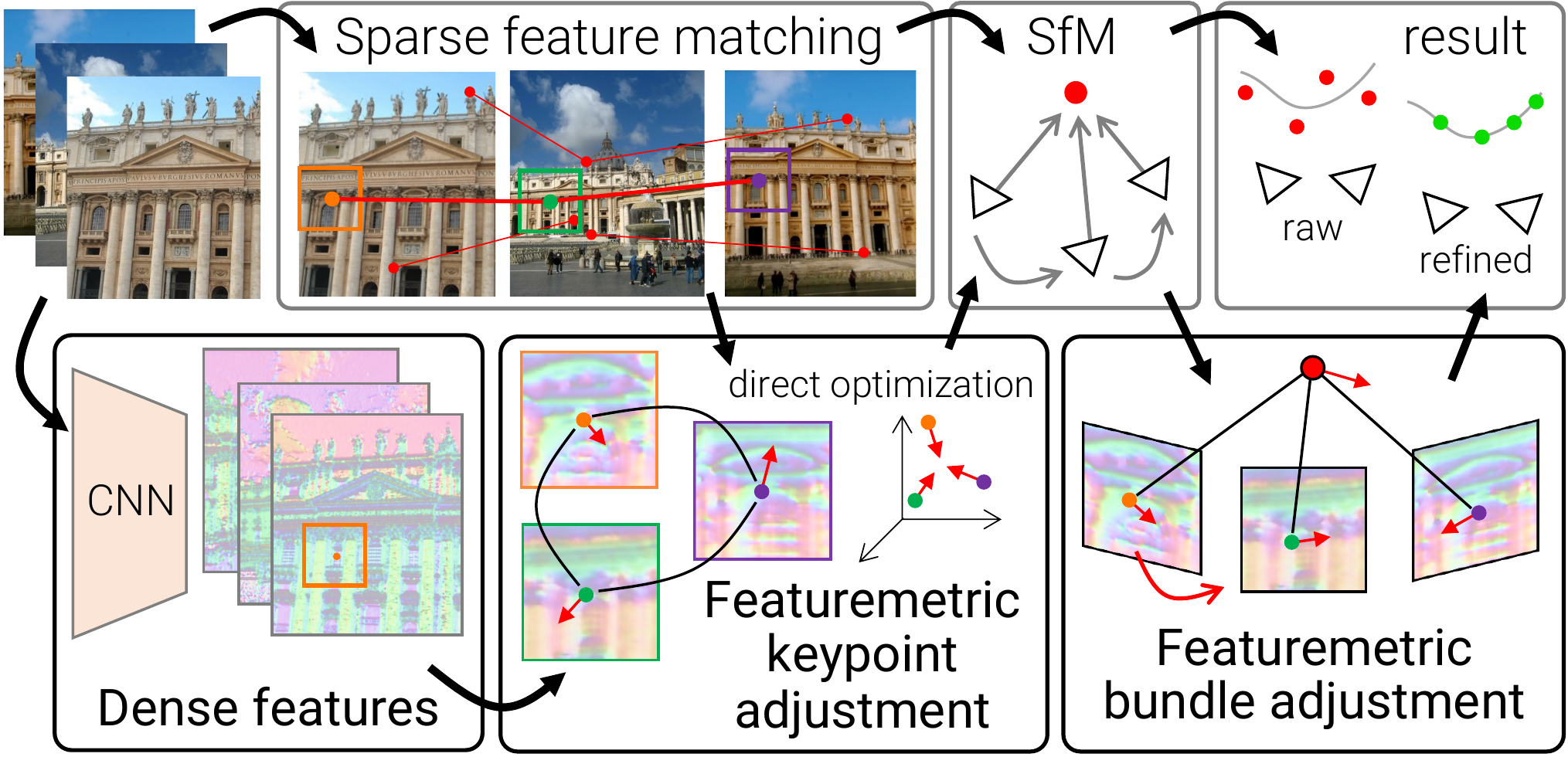}
    \caption{\textbf{Refinement pipeline.}
    Our refinement works on top of any SfM pipeline that is based on local features.
    We perform a two-stage adjustment of keypoints and bundles.
    The approach first refines the 2D keypoints only from tentative matches by optimizing a direct cost over dense feature maps.
    The second stage operates after SfM and refines 3D points and poses with a similar featuremetric cost.
    }%
    \label{fig:pipeline}%
\end{figure}

\section{Related work}

\PAR{Image matching} is at the core of SfM and visual SLAM, which typically rely on sparse local features for their efficiency and robustness. The process i)~detects a small number of interest points, ii)~computes their visual descriptors, iii)~matches them with a nearest neighbor search, and iv)~verifies the matches with two-view epipolar estimation and RANSAC. The correspondences then serve for relative or absolute pose estimation and 3D triangulation. As keypoints are sparse, small inaccuracies in their locations can result in large errors for the estimated geometric quantities.

Differently, dense matching~\cite{liu2010sift, tola2009daisy, UCN2016, Rocco18b, shen2020ransac, GLUNet_Truong_2020, sun2021loftr} considers all pixels in each image, resulting in denser and more accurate correspondences.
It has been successful for constrained settings like optical flow~\cite{ilg2017flownet, Sun2018PWC-Net} or stereo depth estimation~\cite{yao2018mvsnet}, but is not suitable for large-scale SfM due to its high computational cost due to many redundant correspondences.
Several recent works~\cite{taira2018inloc, Rocco20, li20dualrc, ZhouCVPRpatch2pix} improve the matching efficiency by first matching coarsely and subsequently refining correspondences using a local search. 
This is however limited to image pairs and thus cannot create point tracks required by SfM.

Our work combines the best of both paradigms by leveraging dense local information to refine sparse observations. It is inherently amenable to SfM as it can optimize all locations over multiple views in a track simultaneously.

\PAR{Subpixel estimation} is a well-studied problem in correspondence search.
Common approaches either upsample the input images or fit polynomials or Gaussian distributions to local image neighborhoods~\cite{forstner1987fast, huertas1986detection, lowe2004distinctive, scharstein2002taxonomy, hirschmuller2002real}.
With the widespread interest in CNNs for local features, solutions tailored to 2D heatmaps have been recently developed, such as learning fine local sub-heatmaps~\cite{Hu_2019_CVPR} or estimating subpixel corrections with regression~\cite{christiansen2019unsuperpoint, tang2020neural} or the soft-argmax~\cite{ono2018lf, yu2020heatmap}.
Cleaner heatmaps can also arise from aggregating predictions over multiple virtual views using data augmentation~\cite{superpoint}.

Detections or local affine frames can be combined across multiple views with known poses in a least-squares geometric optimization~\cite{triggs1999bundle, Eichhardt_Barath_2019_BMVC}.
Dusmanu~\etal~\cite{dusmanu2020} instead refine keypoints solely based on tentative matches, without assuming known geometry.
This geometric formulation exhibits remarkable robustness, but is based on a local optical flow whose estimation for each correspondence is expensive and approximate.
We unify both keypoint and bundle optimizations into a joint framework that optimizes a featuremetric cost, resulting in more accurate geometries and a more efficient keypoint refinement.

\PAR{Direct alignment} optimizes differences in pixel intensities by implicitly defining correspondences through the motion and geometry.
It therefore does not suffer from geometric noise and is naturally subpixel accurate via image interpolation.
Direct photometric optimization has been successfully applied to optical flow~\cite{LK, LK20years}, visual odometry~\cite{kerl2013dense, engel14eccv, engel2017direct, Czarnowski_2017_ICCV}, SLAM~\cite{alismail2016photometric, Schops_2019_CVPR}, multi-view stereo~(MVS)~\cite{devernay1994computing, delaunoy2014photometric, yao2019recurrent}, and pose refinement~\cite{schops2017multi}.
It generally fails for moderate displacements or appearances changes, and is thus not suitable for large-baseline SfM.
One notable work by Woodford \& Rosten~\cite{woodford2020large} refines dense SfM+MVS models with a robust image normalization.
It focuses on dense mapping with accurate initial poses and moderate appearance changes.
Georgel~\etal~\cite{Georgel2008AUA} instead estimate more accurate relative poses by elegantly combining photometric and geometric costs.
They show that dense information can improve sparse estimation but their approach ignores appearance changes.
Differently, our work improves the entire SfM pipeline starting with tentative matches and addresses larger, challenging changes.

To improve on the weaknesses of photometric optimization, numerous recent works align multi-dimensional image representations.
Examples of this \emph{featuremetric} optimization include frame tracking with handcrafted~\cite{alismail2016robust, park2017icra} or learned descriptors~\cite{lv2019taking, clark2018ls, xu2020deep, von2020gn, von2020lm}, optical flow~\cite{antonakos2015feature, chang2017clkn}, MVS~\cite{yu2020fast}, and dense SfM in small scenes~\cite{tang2018ba}.
Closer to our work, PixLoc~\cite{sarlin21pixloc} learns deep features with a large basin of convergence for wide-baseline pose refinement.
It improves the accuracy of sparse matching but is designed for single images and disregards the scalability to multiple images or large scenes.
Here we extend this paradigm to other steps of SfM and propose an efficient algorithm that scales to thousands of images.
We show that learning task-specific wide-context features is not necessary and demonstrate highly accurate refinements with off-the-shelf features.

In conclusion, our work is the first to apply robust featuremetric optimization to a large-scale sparse reconstruction problem and show significant benefits for visual localization.

\section{Background}
Given $N$ images $\{\*I_i\}$ observing a scene, we are interested in accurately estimating its 3D structure, represented as sparse points $\{\*P_j \in \mathbb{R}^3\}$,  intrinsic parameters $\{\*C_i\}$ of the cameras, and the poses $\{(\*R_i, \*t_i) \in \*{SE}(3)\}$ of the images, represented as rotation matrices and translation vectors.

A typical SfM pipeline performs geometric estimation from correspondences between sparse 2D keypoints $\{\*p_u\}$ observing the same 3D point from different views, collectively called a track.
Association between observations is based on matching local image descriptors $\{\*d_u \in \mathbb{R}^D\}$, but the estimated geometry relies solely on the location of the keypoints, whose accuracy is thus critical.
Keypoints are detected from local image information for each image individually, without considering multiple views simultaneously.
Subsequent steps of the pipeline discover additional information about the scene, such as its its geometry or its multi-view appearance.
Two approaches leverage this information to reduce the detection noise and refine the keypoints.

\PAR{Global refinement:} Bundle adjustment~\cite{triggs1999bundle} is the gold standard for refining structure and poses given initial estimates. It minimizes the total geometric error
\begin{equation}
    E_\mathrm{BA} = \sum_j \sum_{(i, u) \in \mathcal{T}(j)} 
        \left\lVert \Pi\left(\*R_i\*P_j+\*t_i, \*C_i\right) - \*p_u \right\rVert_\gamma \enspace,
    \label{eq:gba}
\end{equation}
where $\mathcal{T}(j)$ is the set the images and keypoints in track~$j$, $\Pi(\cdot)$ projects to the image plane, and $\lVert\cdot\rVert_\gamma$ is a robust norm~\cite{hampel1986robust}.
This formulation implicitly refines the keypoints while ensuring their geometric consistency.
It however ignores the uncertainty of the initial detections and thus requires many observations to reduce the geometric noise.
Operating on an existing reconstruction, it cannot recover observations arising from noisy keypoints that are matched correctly but discarded by the geometric verification.

\PAR{Track refinement:}
To improve the accuracy of the keypoints prior to any geometric 3D estimation, Dusmanu~\etal~\cite{dusmanu2020} optimize their locations over tentative tracks formed by raw, unverified matches.
They exploit the inherent structure of the matching graph to discard incorrect matches without relying on geometric constraints.
Given two-view dense flow fields $\{\*T_{v\rightarrow u}\}$ between the neighborhoods of matching keypoints~$u$ and~$v$, this \emph{keypoint adjustment} optimizes, for each tentative track~$j$, the multi-view cost
\begin{equation}
    E_\mathrm{KA}^j = \sum_{(u, v)\in\mathcal{M}(j)} 
        \left\lVert \*p_v + \*T_{v\rightarrow u}[\*p_v] - \*p_u \right\rVert_\gamma \enspace,
    \label{eq:gka}
\end{equation}
where $\mathcal{M}(i)$ denotes the set of matches that forms the track and $[\cdot]$ is a lookup with subpixel interpolation.
A deep neural network is trained to regress the flow of a single point from two input patches and the flow field is interpolated from a sparse grid.
This dramatically improves the keypoint accuracy, but some errors remain as the regression and interpolation are only approximate.

Both bundle and keypoint adjustments are based on geometric observations, namely keypoint locations and flow, but do not account for their respective uncertainties.
They thus require a large number of observations to average out the geometric noise and their accuracy is in practice limited.

\section{Approach}

Summarizing dense image information into sparse points is necessary to perform global data association and optimization at scale.
However, refining geometry is an inherently local operation, which, we show, can efficiently benefit from locally-dense pixels.
Given constraints provided by coarse but global correspondences or initial 3D geometry, the dense information only needs to be locally accurate and invariant but not globally discriminative.
While SfM typically discards image information as early as possible, we instead exploit it in several steps of the process thanks to direct alignment.
Leveraging the power of deep features, this translates into featuremetric keypoint and bundle adjustments that elegantly integrate into any SfM pipeline by replacing their geometric counterparts. Figure~\ref{fig:pipeline} shows an overview.

We first introduce the featuremetric optimization in Section~\ref{section:featuremetric}. We then describe our formulations of keypoint adjustment, in Section~\ref{section:fm-ka}, and bundle adjustment, in Section~\ref{section:fm-ba}, and analyze their efficiency.

\subsection{Featuremetric optimization}
\label{section:featuremetric}
\PAR{Direct alignment:}
We consider the error between image intensities at two sparse observations: $\*r = \*I_i[\*p_u] - \*I_j[\*p_v]$.
Local image derivatives implicitly define a flow from one point to the other through a gradient descent update:
\begin{equation}
    \*T_{v\rightarrow u}[\*p_v]\ \propto -\frac{\partial \*I_j}{\partial\*p}[\*p_v]^\top\,\*r \enspace.
\end{equation}
This flow can be efficiently computed at any location in a neighborhood around $v$, without approximate interpolation nor descriptor matching.
It naturally emerges from the direct optimization of the photometric error, which can be minimized with second-order methods in the same way as the aforementioned geometric costs.
Unlike the flow regressed from a black-box neural network~\cite{dusmanu2020}, this flow can be made consistent across multiple view by jointly optimizing the cost over all pairs of observations.

\PAR{Learned representation:} 
SfM can handle image collections with unconstrained viewing conditions exhibiting large changes in terms of illumination, resolution, or camera models.
The image representation used should be robust to such changes and ensure an accurate refinement in any condition.
We thus turn to features computed by deep CNNs, which can exhibit high invariance by capturing a large context, yet retain fine local details.
For each image $\*I_i$, we compute a $D$-dimensional, L2-normalized feature map $\*F_i\in\mathbb{R}^{W\times H\times D}$ at identical resolution.
We use the same representations for keypoint and bundle adjustments, requiring a single forward pass per image.
Our experiments show that multiple off-the-shelf dense local descriptors can result in highly accurate refinements.
However, our formulation can also be applied to robust intensity representations, such as the normalized cross-correlation (NCC) over local image patches~\cite{woodford2020large}.

\subsection{Keypoint adjustment}
\label{section:fm-ka}
Once local features are detected, described, and matched, we refine the keypoint locations before geometrically verifying the tentative matches.

\PAR{Track separation:}
Connected components in the matching graph define tentative tracks -- sets of keypoints that are likely to observe the same 3D point, but whose observations have not yet been geometrically verified.
Because a 3D point has a single projection on a given image plane, valid tracks cannot contain multiple keypoints detected in the same image.
We can leverage this property to efficiently prune out most incorrect matches using the track separation algorithm introduced in~\cite{dusmanu2020}.
This speeds up the subsequent optimization and reduces the noise in the estimation.

\PAR{Objective:}
We then adjust the locations of 2D keypoints belonging to the same track $j$ by optimizing its featuremetric consistency along tentative matches with the cost
\begin{equation}
    E_\mathrm{FKA}^j = \sum_{(u, v)\in\mathcal{M}(j)} 
        w_{uv} \left\lVert \*F_{i(u)}[\*p_u] - \*F_{i(v)}[\*p_v] \right\rVert_\gamma,
    \label{eq:fka}
\end{equation}
where $w_{uv}$ is the confidence of the correspondence $(u, v)$, such as the similarity of its local feature descriptors $\*d_u^\top\*d_v$.
This allows the optimization to split tracks connected by weak correspondences, providing robustness to mismatches.
The confidence is not based on the dense features since these are not expected to disambiguate correspondences at the global image level.

\PAR{Efficiency}:
This direct formulation simply compares pre-computed features on sparse points and is thus much more scalable than patch flow regression (Eq.~\ref{eq:gka}), which performs a dense local correlation for each correspondence.
All tracks are optimized independently, which is very fast in practice despite the sheer number of tentative matches.

\PAR{Drift:}
Because of the lack of geometric constraints, the points are free to move anywhere on the underlying 3D surface of the scene.
The featuremetric cost biases the updates towards areas with low spatial feature gradients and with better-defined features.
This can result in a large drift if not accounted for.
Keypoints should however remain repeatable w.r.t.\ unrefined detections to ensure the matchability of new images, such as for visual localization.
It is thus critical to limit the drift, while allowing the refinement of noisier keypoints.
For each track, we freeze the location of the keypoint $\bar{u}$ with highest connectivity, as in~\cite{dusmanu2020}, and constrain the location $\*p_u$ of each keypoint w.r.t.\ to its initial detection $\*p^0_u$, such that $\left\lVert\*p_u - \*p^0_u\right\rVert \leq K$.

Once all tracks are refined, the geometric estimation proceeds, typically using two-view epipolar geometric verification followed by incremental or global SfM.

\begin{table*}[ht]
\centering
\begin{minipage}{0.73\linewidth}
\footnotesize{\setlength\tabcolsep{4.5pt}\
\begin{tabular}{lcccccccccccc}
    \toprule
    \multirow{3}{1.7cm}[-.6em]{SfM features $\drsh$ Refinement}
    & \multicolumn{6}{c}{ETH3D indoor} & \multicolumn{6}{c}{ETH3D outdoor}\\
    \cmidrule(lr){2-7}
    \cmidrule(lr){8-13}
    & \multicolumn{3}{c}{Accuracy (\%)} & \multicolumn{3}{c}{Completeness (\%)}
    & \multicolumn{3}{c}{Accuracy (\%)} & \multicolumn{3}{c}{Completeness (\%)}\\
    \cmidrule(lr){2-4}
    \cmidrule(lr){5-7}
    \cmidrule(lr){8-10}
    \cmidrule(lr){11-13}
    & 1cm & 2cm & 5cm & 1cm & 2cm & 5cm & 1cm & 2cm & 5cm & 1cm & 2cm & 5cm\\
    \midrule
    SIFT~\cite{lowe2004distinctive} & 75.62&85.04&92.45&0.21&0.87&3.61&57.64&71.92&85.23&0.06&0.34&2.45\\
    $\drsh$ Patch Flow & 80.99&89.06&\b{95.06}&0.24&0.97&\b{3.88}&64.79&78.90&90.04&\b{0.08}&0.41&\b{2.76} \\
    $\drsh$ \b{ours} & \b{82.82}&\b{89.77}&94.77&\b{0.25}&\b{0.96}&3.75&\b{68.43}&\b{80.73}&\b{91.28}&\b{0.08}&\b{0.42}&2.75\\
    \midrule
    SuperPoint~\cite{superpoint} &75.76&85.61&93.38&0.59&2.21&8.89&50.45&65.07&80.26&0.10&0.55&3.92\\
    $\drsh$ Patch Flow & 85.77&91.57&95.85&0.72&2.51&\b{9.59}&64.94&77.65&88.86&0.15&0.77&4.93\\
    $\drsh$ \b{ours} & \b{89.33}&\b{93.58}&\b{96.58}&\b{0.74}&\b{2.53}&9.51&\b{71.27}&\b{82.58}&\b{92.08}&\b{0.16}&\b{0.83}&\b{5.06}\\
    \midrule
    D2-Net~\cite{dusmanu2019d2} & 47.18&64.94&83.37&0.47&1.87&7.07&20.87&34.55&56.53&0.03&0.19&1.78\\
    $\drsh$ Patch Flow & 79.10&86.64&93.26&\b{1.45}&\b{4.53}&\b{12.95}&57.34&70.71&84.12&\b{0.21}&\b{1.06}&\b{6.02}\\
    $\drsh$ \b{ours} & \b{82.49}&\b{88.83}&\b{94.35}&1.36&4.13&11.80&\b{65.71}&\b{77.95}&\b{89.22}&\b{0.21}&1.01&5.63\\
    \midrule
    R2D2~\cite{revaud2019r2d2} & 66.30&79.21&90.00&0.53&2.06&8.62&49.32&66.10&83.10&0.11&0.55&3.63\\
    $\drsh$ Patch Flow & 77.94&85.82&92.48&0.66&\b{2.32}&\b{9.07}&64.14&78.10&90.18&\b{0.16}&0.71&\b{4.09}\\
    $\drsh$ \b{ours} & \b{80.67}&\b{87.61}&\b{93.42}&\b{0.67}&2.31&8.95&\b{67.77}&\b{80.85}&\b{91.91}&\b{0.16}&\b{0.73}&\b{4.09}\\
    \bottomrule
\end{tabular}
}
\end{minipage}
\hspace{.5mm}
\begin{minipage}{0.24\linewidth}
\centering
\footnotesize{SuperPoint - raw}
\vspace{.5mm}

\includegraphics[width=\textwidth]{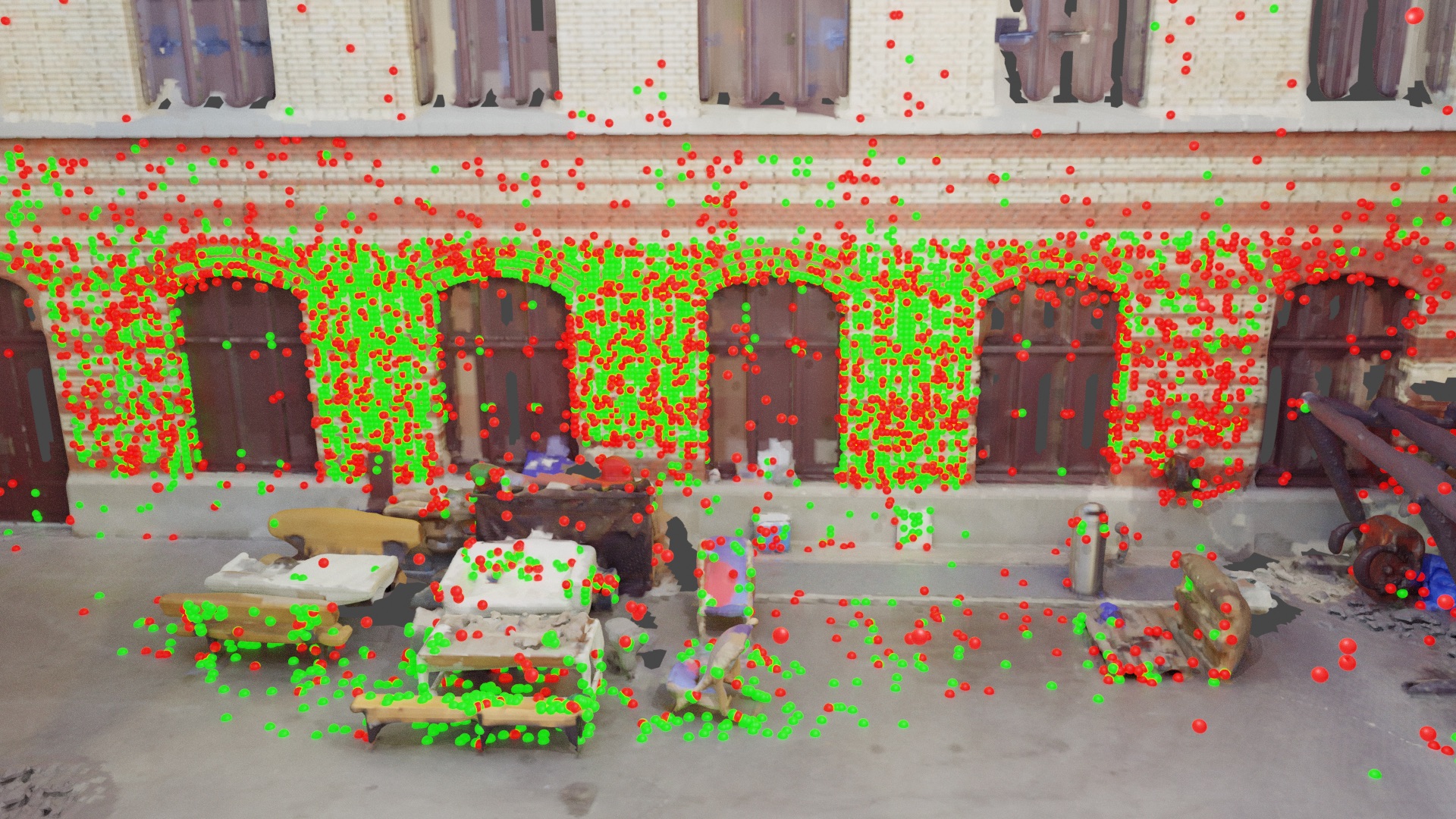}\vspace{2mm}

\footnotesize{SuperPoint - refined}
\vspace{.5mm}

\includegraphics[width=\textwidth]{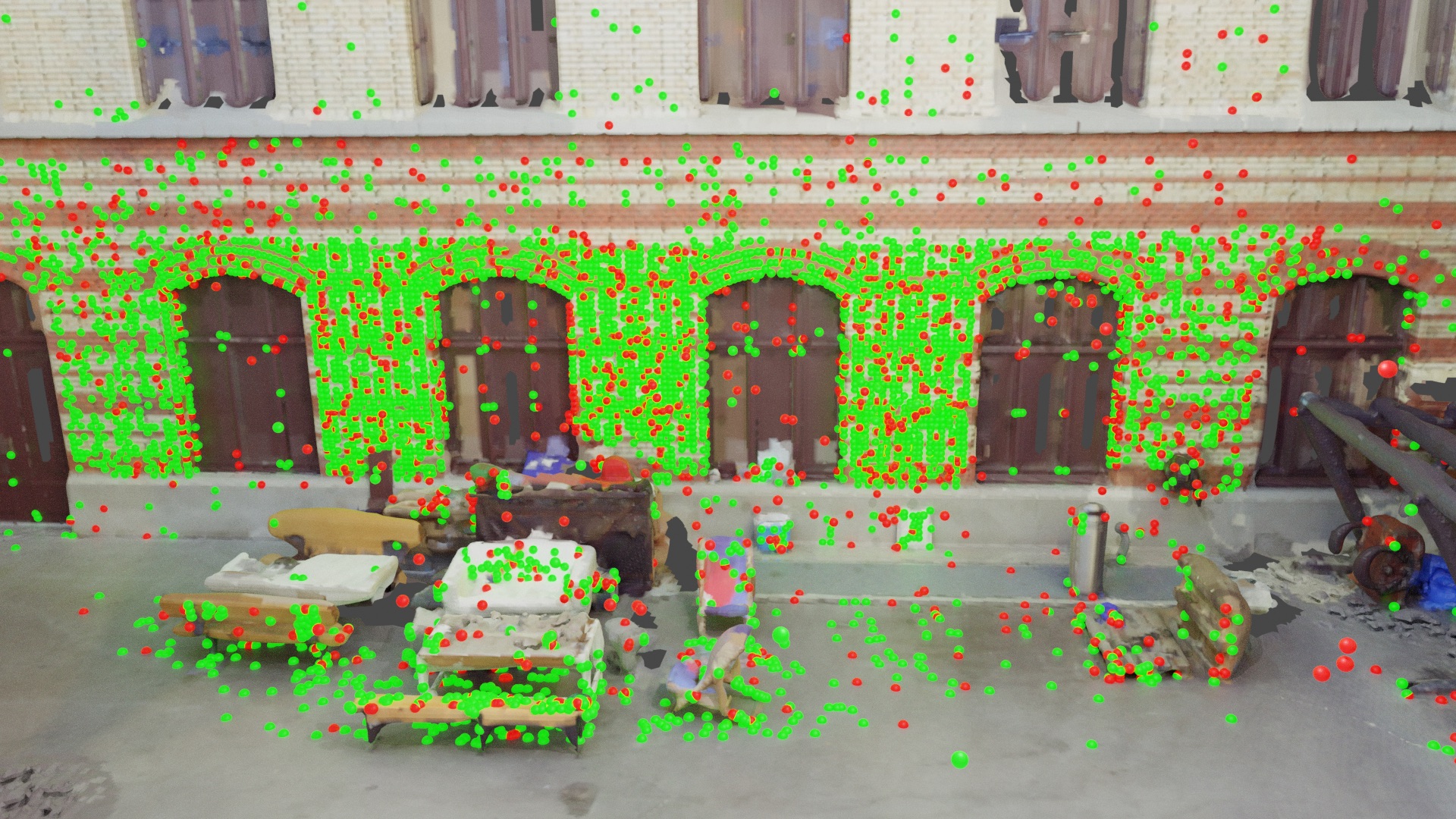}

\green{correct}/\red{incorrect} @ 1cm
\end{minipage}
\vspace{-2mm}
\caption{\textbf{3D sparse triangulation.}
Our refinement yields significantly more accurate and complete point clouds than the common geometric SfM pipeline.
It is more effective than the existing Patch Flow~\cite{dusmanu2020}, especially at 1cm or with SIFT.
}%
\label{tab:eth3d-triangulation}
\end{table*}

\subsection{Bundle adjustment}
\label{section:fm-ba}
The estimated structure and motion can then be refined with a similar featuremetric cost. Here keypoints are implicitly defined by the projections of the 3D points into the 2D image planes, and only poses and 3D points are optimized.

\PAR{Objective:}
We minimize for each track $j$ the error between its observations and a reference appearance $\*f^j$:
\begin{equation}
    E_\mathrm{FBA} = \sum_j \sum_{(i, u) \in \mathcal{T}(j)} 
        \left\lVert \*F_i\left[\Pi\left(\*R_i\*P_j+\*t_i, \*C_i\right)\right] - \*f^j\right\rVert_\gamma \enspace.
    \label{eq:fba}
\end{equation}
The reference is selected at the beginning of the optimization and kept fixed from then on.
This reduces the drift of the points significantly, as also noted in~\cite{alismail2016photometric}, but is more flexible than the common ray-based parametrization~\cite{kerl2013dense, engel2017direct, woodford2020large}.

The reference is defined as the observation closest to the robust mean $\*\mu$ over all initial observations $\*f^j_u$ of the track:
\begin{align}
    \*f^j &= \underset{\*f \in \{\*f^j_u\}}{\mathrm{argmin}} 
        \left\lVert \*\mu^j - \*f \right\rVert \\
    \mathrm{with} \enspace
    \*\mu^j &=  \underset{\*\mu \in \mathbb{R}^D}{\mathrm{argmin}}
        \sum_{\*f \in \{\*f^j_u\}} \left\lVert \*f - \*\mu \right\rVert_\gamma \enspace.
\label{eq:reference}
\end{align}
This ensures robustness to outlier observations and accounts for the unknown topology of the feature space.

\PAR{Efficiency:}
Compared to the keypoint adjustment (Eq.~\ref{eq:fka}), using a reference feature reduces the number of residuals from $\mathcal{O}(N^2)$ to $\mathcal{O}(N)$.
On the other hand, all tracks need to be updated simultaneously because of the interdependency caused by the camera poses.
To accelerate the convergence, we form a reduced camera system based on the Schur complement and use embedded point iterations~\cite{jeong2011pushing}.
The refinement generally converges within a few camera updates.

\subsection{Implementation}
\label{section:implementation}
\PAR{Dense  extractor:}
Our refinement can work with any off-the-shelf CNN that produces feature maps that are locally discriminative.
These should be of the same resolution as the input (stride 1) to enable subpixel accuracy.
The radius of convergence, or context, of such features depends on the amount of noise in the keypoints. 
Most detectors like SIFT have at most a few pixels of error, while others like D2-Net exhibit a much larger detection noise.
In our experiments, we use S2DNet~\cite{germain2020s2dnet} for dense feature extraction, as it computes fine features very efficiently in only 4 convolutions, but also produce, if required, deeper features with a larger context.
These can then be combined into a multi-level optimization scheme~\cite{engel2017direct, von2020gn, sarlin21pixloc} that sequentially refines based on coarse to fine features.
The convergence can thus be adjusted depending on the detector and on the image resolution.
We show in Section~\ref{section:ablation} that other dense features work well too.

\PAR{Optimization:}
The optimization problems of both keypoint and bundle adjustments are solved with the Levenberg-Marquardt~\cite{levenberg1944method} algorithm implemented using Ceres~\cite{ceres-solver}.
Feature maps are stored as collections of 16$\times$16 patches centered around the initial keypoint detections.
We thus constrain points to move at most $K{=}\,$8 pixels.
The feature lookup is implemented as bicubic interpolation.
We use the Cauchy loss $\gamma$ with a scale of 0.25.
The robust mean in Eq.~\ref{eq:reference} is computed with iteratively reweighted least squares~\cite{holland1977robust}.

Simultaneously storing all high-dimensional feature patches incurs high memory requirements during BA.
We dramatically increase its efficiency by exhaustively precomputing patches of feature distances and directly interpolate an approximate cost $\bar{E}_{ij} = \left\lVert\*F_i - \*f^j\right\rVert_\gamma\left[\*p_{ij}\right]$.
To improve the convergence, we store and optimize its spatial derivatives $\nicefrac{\partial \bar{E}_{ij}}{\partial \*p_{ij}}$.
This reduces the residual size from $D$ to 3 with no loss of accuracy.
See \supp~\ref{section:supp:costmap} for more details.

\PAR{Run time and memory:}
S2DNet can extract 3-5 dense feature maps per second and both featuremetric adjustments run in less than 5 minutes for 100 images.
As these features are 128-dimensional, the memory consumption can be a bottleneck.
We believe that much fewer dimensions are actually required for refinement, and retraining a compact feature extractor would improve the efficiency of the optimization.

\section{Experiments}
We evaluate our featuremetric refinement on various SfM tasks with several handcrafted and learned local features and show substantial improvements for all of them.
We first evaluate its accuracy on the tasks of triangulation and camera pose estimation in Sections~\ref{section:triangulation} and~\ref{section:pose}, respectively.
We then assess in Section~\ref{section:phototourism} the impact of the refinement on two-view and multi-view pose estimation for end-to-end reconstruction in challenging conditions. 
Lastly, Section~\ref{section:ablation} analyzes the validity and scalability of our design decisions through an ablation study.

\subsection{3D triangulation}
\label{section:triangulation}

We first evaluate the accuracy of the refined 3D structure given known camera poses and intrinsics.

\PAR{Evaluation:}
We use the ETH3D benchmark~\cite{schops2017multi}, which is composed of 13 indoor and outdoor scenes and provides images with millimeter-accurate camera poses and highly-accurate ground truth dense reconstructions obtained with a laser scanner.
We follow the protocol introduced in~\cite{dusmanu2020}, in which a sparse 3D model is triangulated for each scene using COLMAP~\cite{schoenberger2016sfm} with fixed camera poses and intrinsics.
Following the original benchmark setup, we report the accuracy and completeness of the reconstruction, in \%, as the ratio of triangulated and ground-truth dense points that are within a given distance of each other.

\PAR{Baselines:} 
We evaluate our featuremetric refinement with the hand-crafted local features SIFT~\cite{lowe2004distinctive} and the learned ones SuperPoint~\cite{superpoint}, D2-Net~\cite{dusmanu2019d2}, and R2D2~\cite{revaud2019r2d2}, using the associated publicly available code repositories.
We compare our approach to the geometric optimization of~\cite{dusmanu2020}, referred here as Patch Flow.
We re-compute the numbers provided in the original paper using the code provided by the authors.

\PAR{Results:}
Table~\ref{tab:eth3d-triangulation} shows that our approach results in significantly more accurate and complete 3D reconstructions compared to the traditional geometric SfM.
It is more accurate than Patch Flow, especially at the strict threshold of 1cm, and exhibits similar completeness. 
The improvements are consistent across all local features, both indoors and outdoors.
The gap with Patch Flow is especially large for SIFT, which already detects well-localized keypoints. 
This confirms that our featuremetric optimization better captures low-level image information and yields a finer alignment.
Patch Flow is more complete for larger thresholds as it partly solves a different problem by increasing the keypoint repeatability with its large receptive field, while we focus on their localization.

\begin{table}[t]
\begin{minipage}{0.475\linewidth}
    \flushleft
    \includegraphics[width=0.95\linewidth]{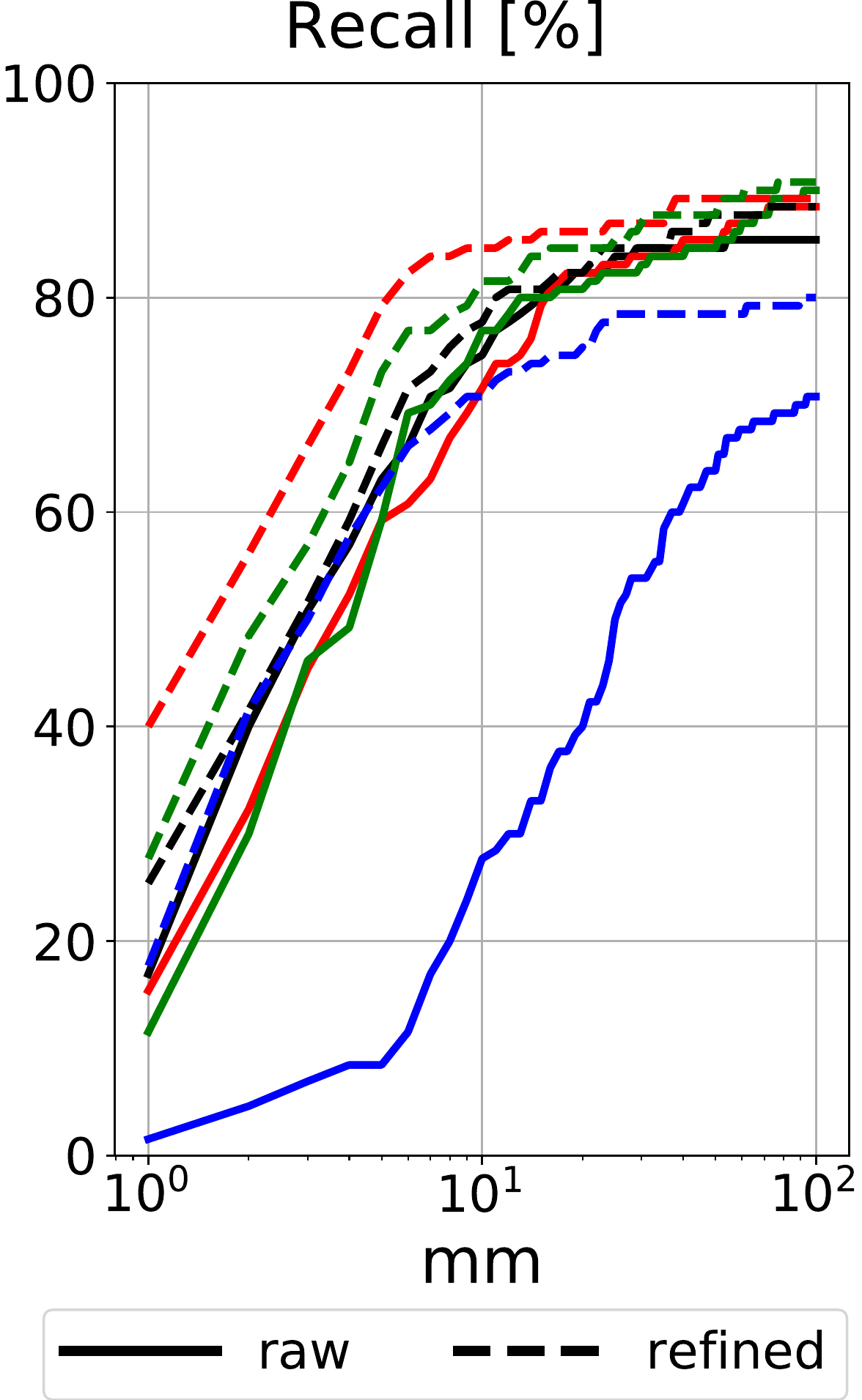}
\end{minipage}%
\begin{minipage}{0.52\linewidth}
    \centering
    \footnotesize{
\setlength\tabcolsep{2.5pt}\
\begin{tabular}{lccc}
    \toprule
    \multirow{2}{1.7cm}[-.3em]{SfM features $\drsh$ Refinement} & \multicolumn{3}{c}{AUC (\%)}\\
    \cmidrule(lr){2-4}
    & 1mm & 1cm & 10cm \\
    \midrule
    {\large $\color{black}\bullet$} SIFT & 16.92 & 56.08 & 81.65 \\
    $\drsh$ Patch Flow & 14.62 & 52.69 & 81.69 \\
    $\drsh$ \b{ours} & \b{25.38} & \b{60.22} & \b{84.07} \\
    \midrule
    {\large $\color{red}\bullet$} SuperPoint & 15.38 & 51.20 & 82.33 \\
    $\drsh$ Patch Flow & 28.46 & 63.99 & 86.79 \\
    $\drsh$ \b{ours} & \b{40.00} & \b{71.97} & \b{86.86} \\
    \midrule
    {\large $\color{blue}\bullet$} D2-Net & \01.54 & 12.16 & 56.10 \\
    $\drsh$ Patch Flow & 16.92 & 54.70 & 75.16 \\
    $\drsh$ \b{ours} & \b{17.69} & \b{55.03} & \b{76.26} \\
    \midrule
    {\large $\color{green!40!black}\bullet$} R2D2 & 11.53 & 52.88 & 82.69 \\
    $\drsh$ Patch Flow & 25.38 & 61.42 & 84.14 \\
    $\drsh$ \b{ours} & \b{27.69} & \b{63.86} & \b{86.13} \\
    \bottomrule
\end{tabular}}
\end{minipage}
\vspace{-1mm}
\caption{\textbf{Camera pose estimation.}
We plot the cumulative translation error and report its AUC. 
Our refinement improves the accuracy of the query camera poses for all local features, even when for SIFT, whose detections are already well-localized.
It is generally more accurate than Patch Flow.
}%
\label{tab:eth3d-localization}
\end{table}

\begin{table}[t]
\centering
\footnotesize{\setlength\tabcolsep{2.7pt}\
\begin{tabular}{lccccc}
    \toprule
    \multirow{3}{1.7cm}[-.3em]{SfM features (\# keypoints) $\drsh$ Refinement}
    & \multicolumn{2}{c}{Task 1: Stereo} & \multicolumn{3}{c}{Task 2: Multiview}\\
    
    \cmidrule(lr){2-3}
    \cmidrule(lr){4-6}
    &\multicolumn{2}{c}{AUC@K\degree} & \multicolumn{3}{c}{AUC@5\degree@$N$}\\
    \cmidrule(lr){2-3}
    \cmidrule(lr){4-6}
    & 5\degree & 10\degree & 5 & 10 & 25\\
    \midrule
    SuperPoint+SuperGlue (2k) & 58.78 & 71.01 & 63.02 & 77.36 & 86.76\\
    $\drsh$ \textbf{ours} & \b{65.89} & \b{76.51} & \b{68.87} & \b{82.09} & \b{89.73}\\
    \midrule
    SIFT (2k) & 38.09 & 48.05 & 25.12 & 50.82 & 77.28\\
    $\drsh$ \textbf{ours} & \b{40.59} & \b{50.87} & \b{28.01} & \b{53.59} & \b{79.49}\\
    \midrule
    D2-Net (4k) & 16.83 & 22.40 & 16.52 & 33.07 & 49.35\\
    $\drsh$ \textbf{ours} & \b{25.89} & \b{33.32} & \b{21.33} & \b{40.69} & \b{57.93}\\
    \bottomrule
\end{tabular}}
\vspace{-1mm}
\caption{\textbf{End-to-end SfM.}
The proposed refinement improves the accuracy of poses estimated by epipolar geometry (stereo) or a complete SfM pipeline (multiview) with crowd-sourced imagery.
Improvements are substantial for both standard (SIFT) and recent (SuperGlue) matching configurations, especially when few images $N$ observe the scene.
}%
\label{tab:phototourism}
\end{table}

\subsection{Camera pose estimation}
\label{section:pose}
We now evaluate the impact of our refinement on the task of camera pose estimation from a single image.

\PAR{Evaluation:}
We again follow the setup of~\cite{dusmanu2020} based on the ETH3D benchmark.
For each scene, 10 images are randomly selected as queries.
For each of them, the remaining images, excluding the 2 most covisible ones, are used to triangulate a sparse 3D partial model.
Each query is then matched against its corresponding partial model and the resulting 2D-3D matches serve to estimate its absolute pose using LO-RANSAC+PnP~\cite{chum2003locally} followed by geometric refinement.
We compare the 130 estimated query poses to their ground truth and report the area under the cumulative translation error curve~(AUC) up to 1mm, 1cm, and 10cm.

\PAR{Baselines:} 
Patch Flow performs multi-view optimization over each partial model independently as well as over the matches between each query and its partial model.
Similarly, we first refine each partial model as in Section~\ref{section:triangulation}.
We then adjust the query keypoints using its tentative matches, estimate an initial pose, and refine it with featuremetric BA.

\PAR{Results:}
The AUC and its cumulative plot are shown in Table~\ref{tab:eth3d-localization}.
Our refinement substantially improves the localization accuracy for all local features, including SIFT, for which Patch Flow does not show any benefit.
At all error thresholds, featuremetric optimization is consistently more accurate than its geometric counterparts.
The accuracy of SuperPoint is raised far higher than other detectors, despite the high sparsity of the 3D models that it produces. 
This shows how more accurate keypoint detections can result in much more accurate visual localization.

\subsection{End-to-end Structure-from-Motion}
\label{section:phototourism}
While the previous experiments precisely quantify the accuracy of the refinement, they do not contain any variations of appearance or camera models.
We thus turn to crowd-sourced imagery and evaluate the benefits of our featuremetric optimization in an end-to-end reconstruction pipeline.

\PAR{Evaluation:}
We use the data, protocol, and code of the 2020 Image Matching Challenge~\cite{Jin2020, imwchallenge2020}.
It is based on large collections of crowd-sourced images depicting popular landmarks around the world.
Pseudo ground truth poses are obtained with SfM~\cite{schoenberger2016sfm} and used for two tasks.
The stereo task evaluates relative poses estimated from image pairs by decomposing their epipolar geometry.
This is a critical step of global SfM as it initializes its global optimization.
The multiview task runs incremental SfM for small subsets of images, making the SfM problem much harder, and evaluates the final relative poses within each subset.
For each task, we report the AUC of the pose error at the threshold of 5\degree, where the pose error is the maximum of the angular errors in rotation and translation. 
As the evaluation server accepts at most correspondences, we cannot evaluate our method using the test data.
We instead test on a subset of the publicly available validation scenes, and tune the RANSAC and matching parameters on the remaining scenes. More details on this setup are provided in the~\supp.

\PAR{Baselines:}
We evaluate our refinement in combination with SIFT~\cite{lowe2004distinctive}, D2-Net~\cite{dusmanu2019d2}, and SuperPoint+SuperGlue~\cite{superpoint, sarlin2020superglue}.
We limit the number of detected keypoints to 2k for computational reasons, but increase this number to 4k for D2-Net as it otherwise performs poorly.
In the stereo task, we adjust the keypoints using the entire exhaustive tentative match graph (4950 pairs per scene).
We use LO-DEGENSAC~\cite{Chum2005, chum2003locally} for match verification, the ratio test for SIFT, and the mutual check for SIFT and D2-Net.
In the multiview task, we adjust keypoints for each subset independently, considering only the matches between images in the subset, and run our bundle adjustment after SfM.

\PAR{Results:} Table~\ref{tab:phototourism} summarizes the results.
For stereo, our featuremetric keypoint adjustment significantly improves the accuracy of the two-view epipolar geometries across all local features and despite the challenging conditions.
In multiview setting, it also improves the accuracy of the SfM poses, especially for small sets of images.
Featuremetric optimization is particularly effective in this situation, as geometric optimization cannot fully suppress the detection noise due to the small number of observations.
We visualize tracks of a 5-image reconstruction in Figure~\ref{fig:qualitative} and highlight the accuracy of the refined SfM model.

\begin{table}[t]
\centering
\footnotesize{\setlength\tabcolsep{3.2pt}\
\begin{tabular}{llcccccc}
    \toprule
    &\multirow{2}{1.7cm}[-.3em]{SuperPoint $\drsh$~Refinement}
    & \multicolumn{2}{c}{\red{Acc.\ (\%)}} & \multicolumn{2}{c}{\red{Compl.\ (\%)}} & \multirow{2}{0.8cm}[-.3em]{\red{track length}} & \blue{AUC}\\
    \cmidrule(lr){3-4}
    \cmidrule(lr){5-6}
    \cmidrule(lr){8-8}
    && 1cm & 2cm & 1cm & 2cm & & 1cm\\
    \midrule
    \multirow{5}{*}{\begin{sideways}KA vs.\ BA\end{sideways}}
    & unrefined & 18.42 & 32.23 & 0.06 & 0.49 & 4.17 & 51.20 \\
    &$\drsh$ Patch Flow~\cite{dusmanu2020} & 37.00 & 55.18 & 0.15 & 0.93 & \b{5.24} & 63.53 \\
    &$\drsh$ F-KA & 36.85 & 54.48 & 0.15 & 0.90 & 5.02 & 69.84 \\
    &$\drsh$ F-BA & 43.65 & 62.44 & 0.18 & 1.06 & 4.17  & 67.61 \\
    &$\drsh$ \b{F-KA+BA (full)} & \b{46.46} & \b{65.41} & \b{0.19} & \b{1.14} & 5.02 & \b{71.97} \\
    \midrule
    \multirow{3}{*}{\begin{sideways}bonus\end{sideways}}
    &w/ F-BA drift & 47.93 & 66.52 & 0.20 & 1.17 & 5.02 & 64.51 \\
    &Patch Flow + F-BA & 46.30 & 65.22 & 0.19 & 1.13 & 5.24 & - \\
    &higher resolution & 47.67 & 65.39 & 0.21 & 1.21 & 5.12 & - \\
    \midrule
    \multirow{4}{*}{\begin{sideways}dense feats\end{sideways}}
    & photometric BA~\cite{woodford2020large}& 28.43 & 45.87  & 0.11 & 0.72 & 4.17 & - \\
    & VGG-16 ImageNet & 36.86 & 54.99 & 0.15 & 0.90 & 4.61 & - \\
    & DSIFT~\cite{liu2010sift}& 38.78 & 56.46  & 0.16 & 0.96 & 4.73 & - \\
    &PixLoc~\cite{sarlin21pixloc} & 29.49 & 46.60 & 0.12 & 0.74 & 4.48 & - \\
    \bottomrule
\end{tabular}}
\vspace{-1mm}
\caption{\textbf{Ablation study on ETH3D.}
i)~Featuremetric keypoint and bundle adjustments (KA and BA) both largely improve the \red{triangulation} and \blue{localization} accuracy. Patch Flow produces a longer track length because of its larger receptive field but is less accurate.
ii)~Letting the BA drift by updating reference features or increasing the image resolution both improve the triangulation, at the expense of poorer localization and increased run time, respectively.
iii)~Different image representations are better than the unrefined detections but S2DNet (our default) works best.
}
\label{tab:eth3d-ablation}
\end{table}

\begin{figure}[ht]
    \centering
    \begin{minipage}{\linewidth}
        \includegraphics[width=\linewidth]{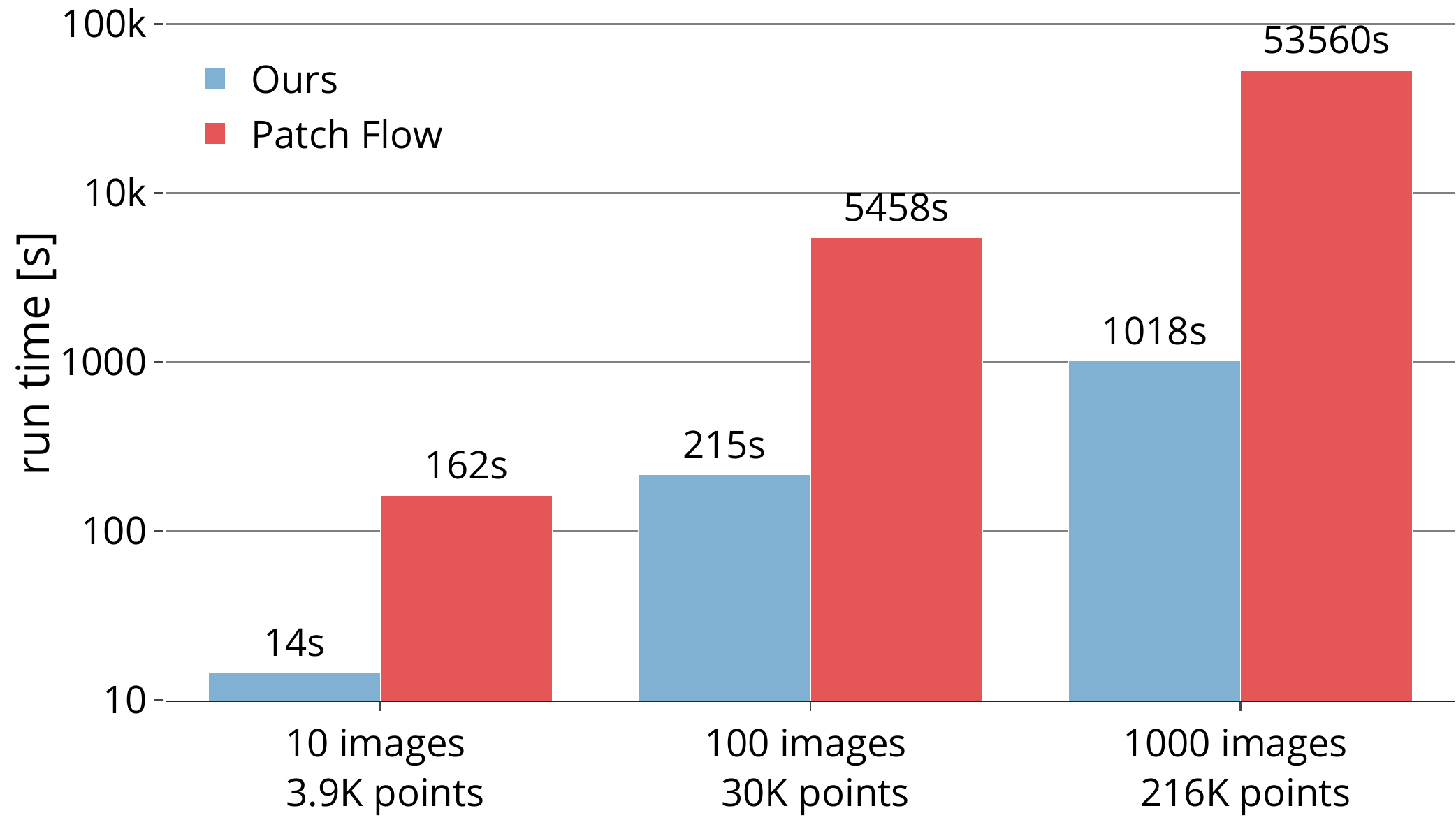}
    \end{minipage}
    \vspace{3mm}
    
    \begin{minipage}{\linewidth}
        \includegraphics[width=\linewidth]{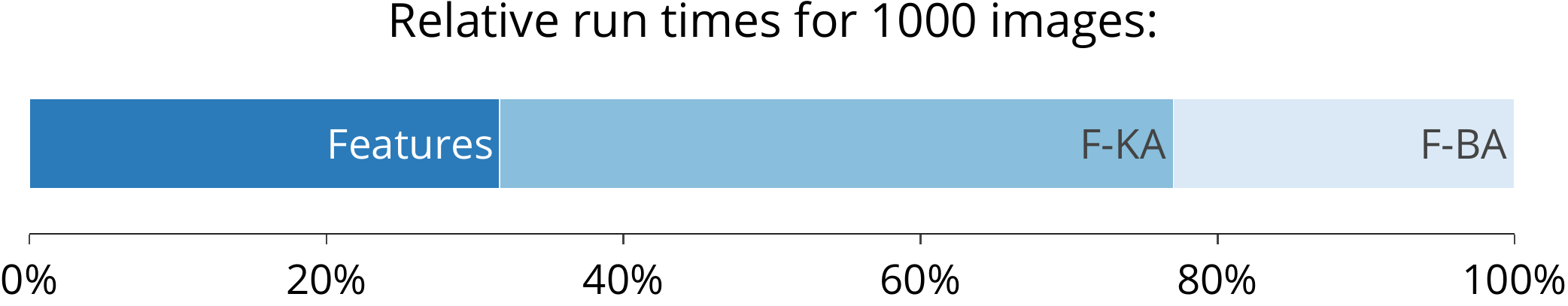}
    \end{minipage}%
    \vspace{-1mm}
    \caption{\textbf{Run-times.}
    We show the duration, in logarithmic scale, of the refinement for varying numbers of images.
    Our refinement is more than ten times faster than Patch Flow~\cite{dusmanu2020}, whose run-time is dominated by the computation of the pairwise flow, which scales quadratically.
    Thanks to our precomputed cost patches, the featuremetric BA is fast.
    The KA amounts for the majority of the refinement time.
    }%
    \label{fig:run-times}%
\end{figure}

\begin{figure*}[t]
    \centering
    \input{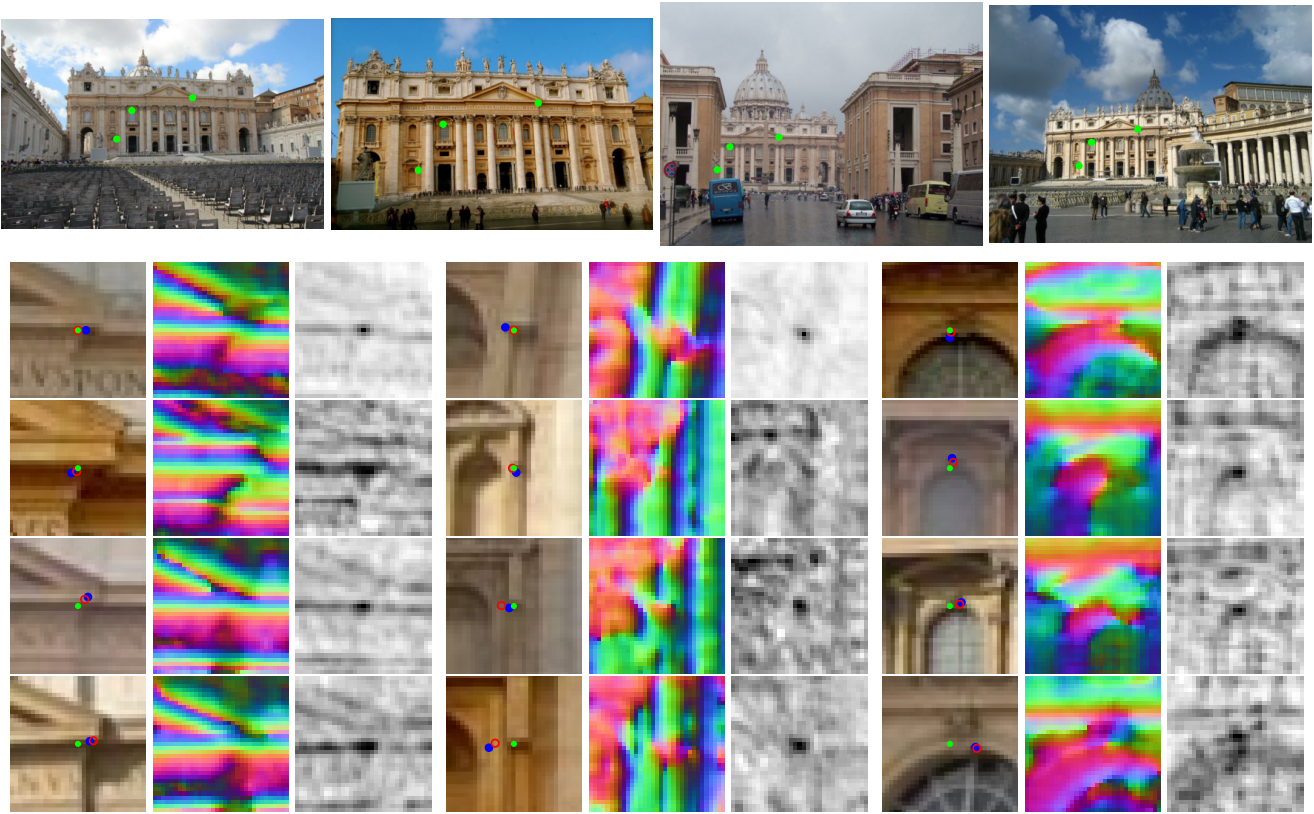}
    \vspace{-2mm}
    \caption{\textbf{Refined SfM tracks.}
    We show patches centered around reprojections of 3x 3D points observed in 4 images of the \emph{St. Peter's Square} scene.
    Deep features and their correlation maps with a reference are robust to scale or illumination changes, yet preserve local details required for fine alignment.
    Points refined with our approach (in \green{green}) are consistent across multiple views while those of a standard SfM pipeline (in \red{red}) are misaligned because the initial keypoint detections (in \blue{blue}) are noisy.
    }%
    \label{fig:qualitative}%
\end{figure*}

\subsection{Additional insights}
\label{section:ablation}

\PAR{Ablation study:}
Table~\ref{tab:eth3d-ablation} shows the performance of several variants of our featuremetric optimization on ETH3D in terms of triangulation~(scene \emph{Facade} only) and localization~(all scenes). We compare both types of adjustments, minor tweaks, and different image representations, including NCC-normalized intensity patches with fronto-parallel warping. Our final configuration, based on on the dense features of S2DNet~\cite{germain2020s2dnet}, performs best across all metrics. We will now show that it is also fairly efficient.

\PAR{Scalability:} 
We run SfM on subsets of images of the Aachen Day-Night dataset \cite{sattler2018benchmarking, sattler2012image, Zhang2020IJCV}.
Figure~\ref{fig:run-times} shows the run times of the refinement for subsets of 10, 100 and 1000 images.
The featuremetric refinement is an order of magnitude faster than Patch-Flow~\cite{dusmanu2020}.
Precomputing distance maps reduces the peak memory requirement of the bundle adjustment from 80 GB to less than 10GB for 1000 images.
As storing feature maps only requires 50 GB of disk space, this refinement can easily run on a desktop PC.
We thus refined the entire Aachen Day-Night v1.1 model, composed of 7k images, in less than 2 hours.
Scene partitioning~\cite{schoenberger2016sfm} could further reduce the peak memory.
See \supp~\ref{section:supp:details} for more details.

\section{Conclusion}
In this paper we argue that the recipe for accurate large-scale Structure-from-Motion is to perform an initial coarse estimation using sparse local features, which are by necessity globally-discriminative, followed by a refinement using locally-accurate dense features. Since the dense feature only need to be locally-discriminative, they can afford to capture much lower-level texture, leading to more accurate correspondences. Through extensive experiments we show that this results in more accurate camera poses and structure; in challenging conditions and for different local features.

While we optimize against dense feature maps, we keep the sparse scene representation of SfM.
This ensures not only that the approach is scalable but also that the resulting 3D model is compatible with downstream applications, \eg~mapping for visual localization.
Since our refinement works well even with few observations, as it does not need to average out the keypoint detection noise, it has the potential to achieve more accurate results using fewer images.

We thus believe that our approach can have a large impact in the localization community as it can improve the accuracy of the ground truth poses of standard benchmark datasets, of which many are currently saturated.
Since this refinement is less sensitive to under-sampling, it enables benchmarking for crowd-sourced scenarios beyond densely-photographed tourism landmarks.

{\footnotesize%
\PAR{Acknowledgements:}
    The authors thank Mihai Dusmanu, Rémi Pautrat, Marcel Geppert, and the anonymous reviewers for their thoughtful comments.
Paul-Edouard Sarlin was supported by gift funding from Huawei, and Viktor Larsson by an ETH~Zurich Postdoctoral Fellowship.
\par
}

\fi 

\ifproceedings
    \ifsupponly
        \appendix

\ifproceedings
\pagestyle{plain}
\begin{strip}
\begin{center}
    \ifsupponly\vspace{-1.4cm}\fi
    {\Large \bf Pixel-Perfect Structure-from-Motion with Featuremetric Refinement\par}
    \vspace{0.5cm}
    {
        \large
        Philipp Lindenberger$^{1}$\printfnsymbol{1}\hspace{.12in}
        Paul-Edouard Sarlin$^{2}$\printfnsymbol{1}\hspace{.12in}
        Viktor Larsson$^{2}$\hspace{.12in}
        Marc Pollefeys$^{2,3}$
        \vspace{0.3cm}\\
        Departments of $^{1}$Mathematics and $^{2}$Computer Science, ETH Zurich\hspace{.2in}
        $^{3}$Microsoft
    }
\end{center}
\end{strip}
\iccvrulercount 0\relax
\setcounter{page}{1}
\fi

\section*{\supp}
\ifproceedings
In the following pages, we present additional details on the experiments conducted in the main paper.
We provide in Section~\ref{section:supp:eth3d} additional detailed results for the tasks of triangulation and camera pose estimation on the ETH3D dataset.
In Section~\ref{section:supp:parameters}, we analyze how several parameters can impact the accuracy and the computational requirements of the refinement.
Section~\ref{section:supp:costmap} describes our efficient cost approximation.
Finally, in Section~\ref{section:supp:details} we provide additional implementation details on the experiments presented in the main paper.
\fi

\section{Additional results on ETH3D}
\label{section:supp:eth3d}
\subsection{Triangulation}
We refine the triangulation of SuperPoint~\cite{superpoint} keypoints for the ETH3D \emph{Courtyard} scene and show in
Figure~\ref{fig:errors-eth3d} the distribution of triangulation errors for points observed by different numbers of images (track length).
Our featuremetric refinement provides the largest improvement for points with low track length, for which the estimates of the traditional geometric BA are dominated by the noise of the keypoint detection.
For larger track lengths, the refined point cloud has an accuracy close to the Faro Focus X 330 laser scanner from which the ground truth is computed.

We show in Figure~\ref{fig:triangulation:courtyard} the raw and refined point clouds for SuperPoint and D2-Net. The benefits of our refinement are easily visible in 3D. Planar walls exhibit fewer noisy keypoints and the refined point clouds are more complete.

\subsection{Camera pose estimation}
\label{subection:supp:camerapose}
We analyze in Table~\ref{tab:eth3d-ablation_localization} how the different kinds of adjustments impact the accuracy of camera localization.
The full method presented in the main paper first refines the 3D SfM model with featuremetric keypoint and bundle adjustments.
It then refines each keypoint in the query image using its tentative 2D-3D correspondences by minimizing the featuremetric error between its observation in the query and the most similar observation of the respective 3D points.
Refining the query keypoints before RANSAC increases the number of inlier matches and stabilizes the pose estimation in challenging scenarios where few 3D points are matches.

Once an initial pose is estimated with PnP+RANSAC, we refine it via a small featuremetric bundle adjustment over the inlier correspondences.
This optimizes each query keypoint against the closest descriptor within the matched track.
As opposed to refining each query keypoint against all observations in the matched track, this has the benefit of scaling linearly in the number of query keypoints and yields a similar accuracy.

\begin{figure}[t]
    \centering
    \includegraphics[width=1.0\linewidth]{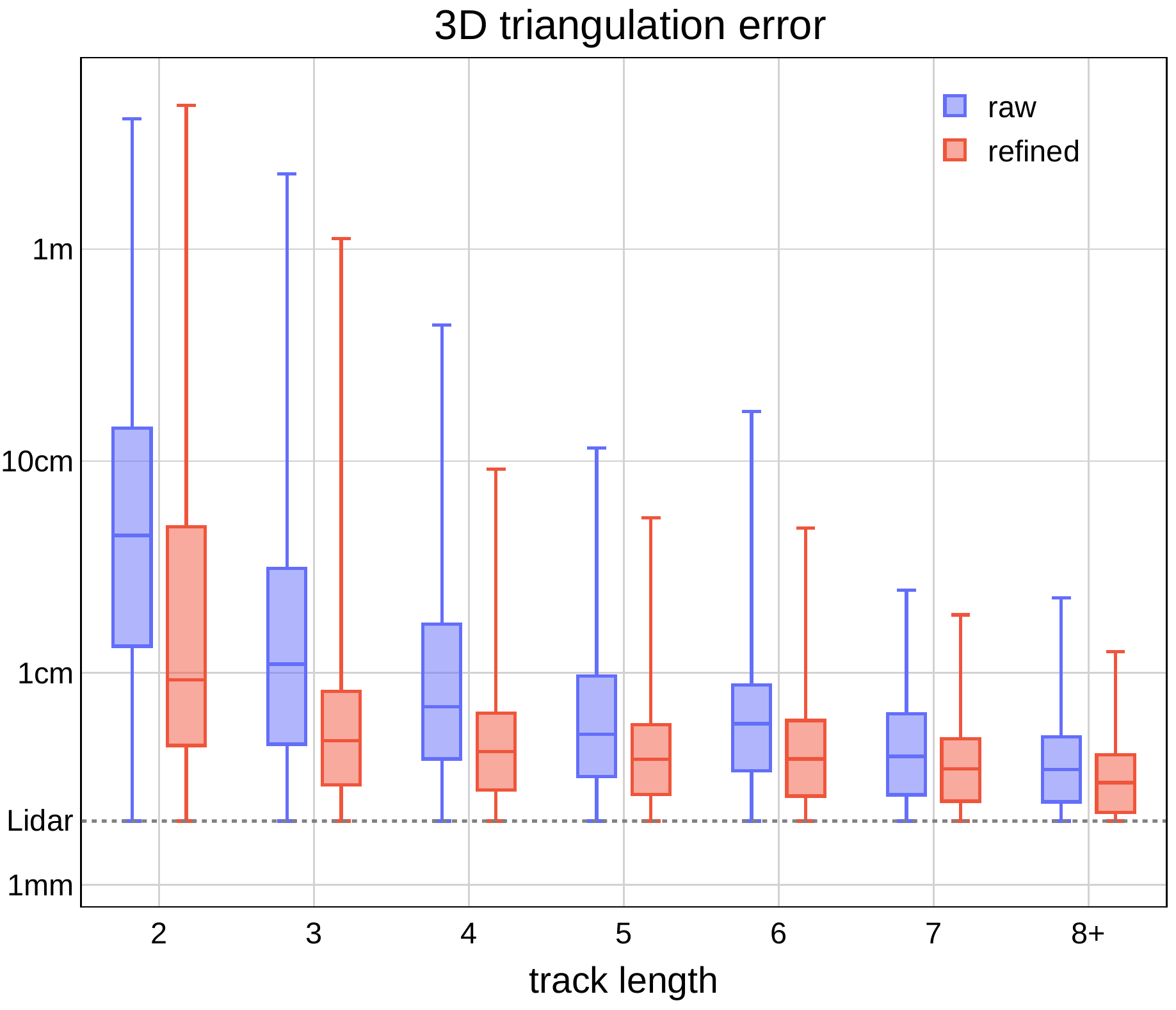}
    \caption{\textbf{Triangulation errors vs.\ track length.}
    The initial, unrefined output, based on geometric BA, exhibits high errors for 3D points that are observed by few images (low track length).
    Our refinement significantly reduces these errors and brings the accuracy of the sparse point cloud close to the ground truth acquired by Lidar (2mm accuracy).
    }
    \label{fig:errors-eth3d}%
\end{figure}
\begin{table}[t]
\centering
\footnotesize{\setlength\tabcolsep{2.5pt}\
\begin{tabular}{lccccccc}
    \toprule
    \multirow{2}{1.7cm}[-.3em]{SuperPoint $\drsh$~Refinement} & KA & BA & qKA & qBA & \multicolumn{3}{c}{AUC (\%)}\\
    \cmidrule(lr){6-8}
    &&&&& 1mm & 1cm & 10cm \\
    \midrule
    unrefined &&&&& 15.38 & 51.20 & 82.33 \\
    $\drsh$ refined &\checkmark&&&& 16.15 & 53.34 & 82.49 \\
    $\drsh$ refined &\checkmark&\checkmark&&& 16.92 & 54.71 & 84.08 \\
    $\drsh$ refined &\checkmark&\checkmark&\checkmark&& 38.46 & 70.44 & 85.28 \\
    $\drsh$ \b{refined (full)} &\checkmark&\checkmark&\checkmark&\checkmark& \b{40.00} & \b{71.97} & \b{86.86} \\
    \midrule
    $\drsh$ Patch Flow &\checkmark&&\checkmark& & 28.46 & 63.04 & 86.65 \\
    \midrule
\end{tabular}}
\caption{\textbf{Ablation study for pose estimation.}
The accuracy of the camera pose is improved by refining the map (KA and BA) and by refining the query keypoints before (qKA) and after (qBA) pose estimation.
The largest improvement is brought by qKA. It increases the number of inlier matches and the likelihood of finding the correct pose with RANSAC.
}
\label{tab:eth3d-ablation_localization}
\end{table}

\begin{figure}[t]
    \centering
    \includegraphics[width=1.0\linewidth]{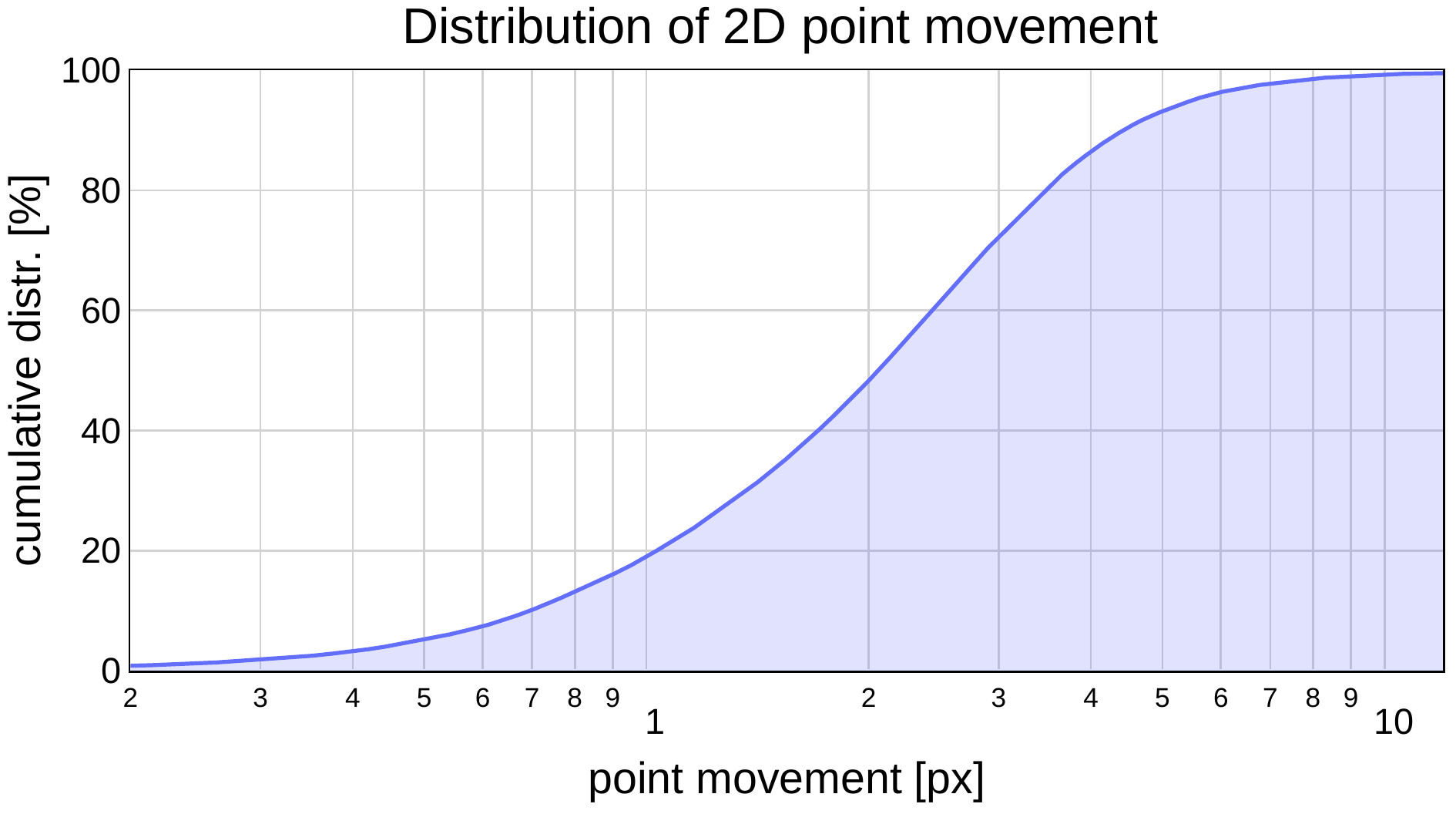}
    \caption{\textbf{Distribution of point movements.}
    We show the cumulative distribution of the distance traveled by the 2D keypoints during the featuremetric refinement of SuperPoint with KA and BA.
    60\% of the points move by fewer than 2 pixels and 99\% remain within 8 pixels of the initial detections.}
    \label{fig:movements-eth3d}%
\end{figure}

\begin{figure}[t]
    \centering
    \includegraphics[width=1.0\linewidth]{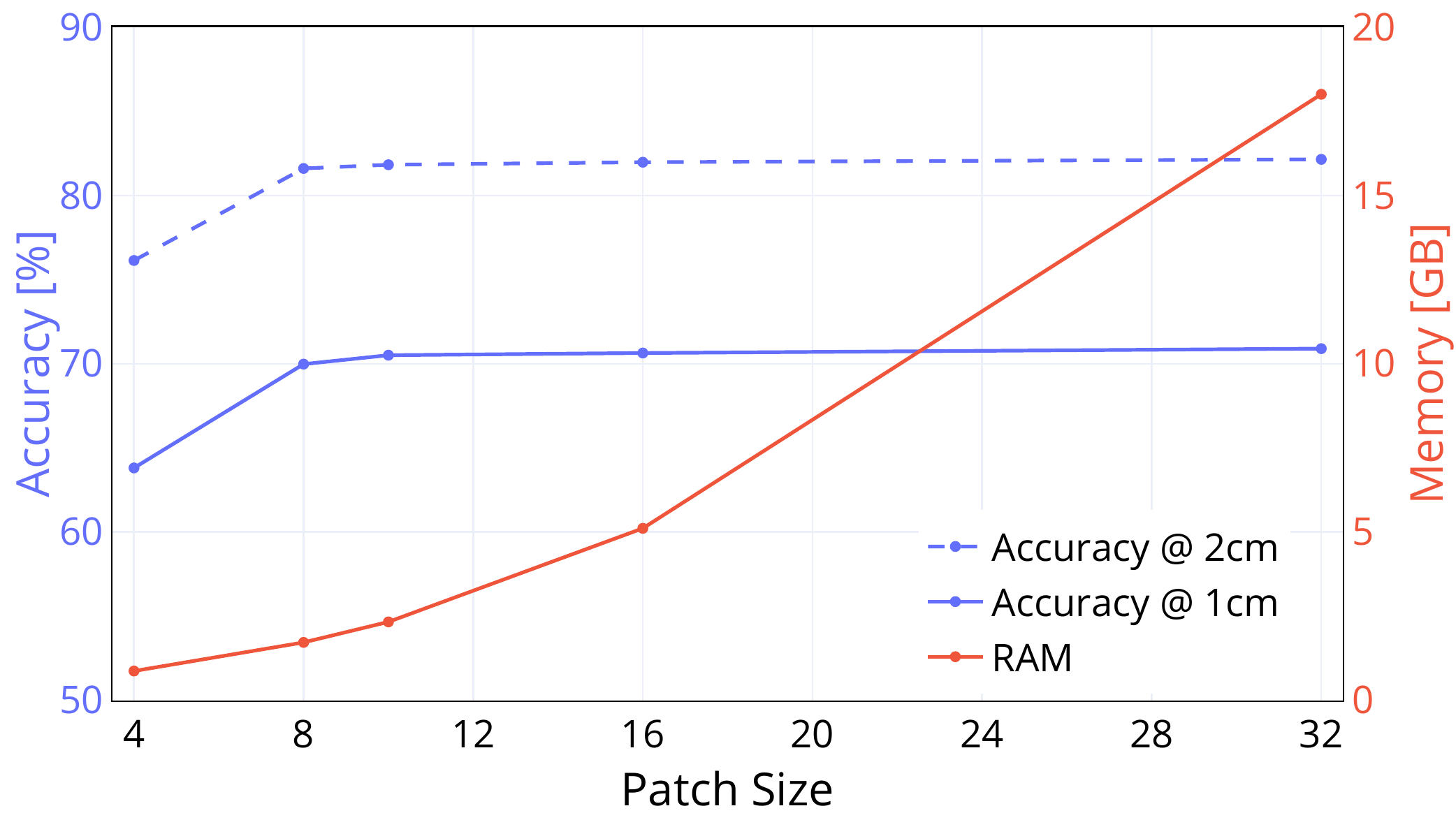}
    \caption{\textbf{Impact of the patch size.}
    Smaller patches for each observation significantly reduce memory requirements but can impair the accuracy of the refinement.
    Patches of size 10$\times$10 offer a good trade-off with high accuracy and moderate memory consumption.
    }
    \label{fig:patch-size}%
\end{figure}

\section{Impact of various parameters}
\label{section:supp:parameters}
\subsection{Patch size}
Figure~\ref{fig:movements-eth3d} shows how much our refinement displaces the detected keypoints during the triangulation of SuperPoint on \emph{Courtyard} using dense features extracted from 1600x1066-pixel images.
When using full feature maps without any constraints in keypoint adjustment, most points are moved by more than 1 pixel, but most often by less than 8 pixels.
This confirms that storing the feature maps as 16$\times$16 patches is sufficient and rather conservative.

We show in Figure~\ref{fig:patch-size} the accuracy of the triangulation for various patch sizes.
Smaller 10$\times$10 patches achieve sufficient accuracys and require significantly less memory.

\begin{figure}[t]
    \centering
    \includegraphics[width=1.0\linewidth]{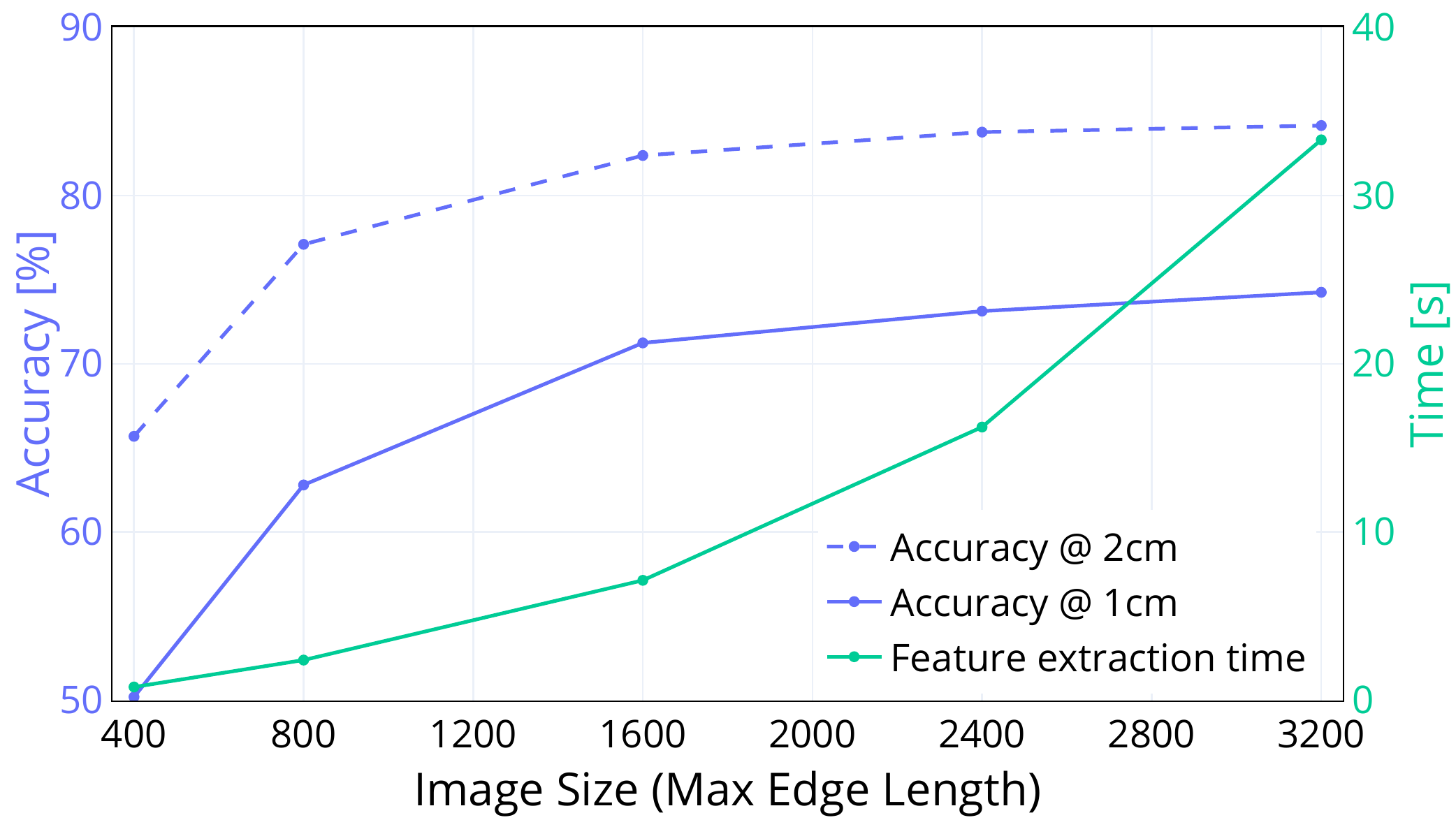}
    \caption{\textbf{Impact of the image resolution.}
    Increasing the image resolution increases the accuracy, but at the cost of longer feature extraction time and higher VRAM requirements.
    For all experiments on ETH3D, we used a maximum edge length of 1600px, which is very close to saturating the accuracy while providing low run times.}
    \label{fig:image-resolution}%
\end{figure}

\begin{figure}[t]
    \centering
    \includegraphics[width=1.0\linewidth]{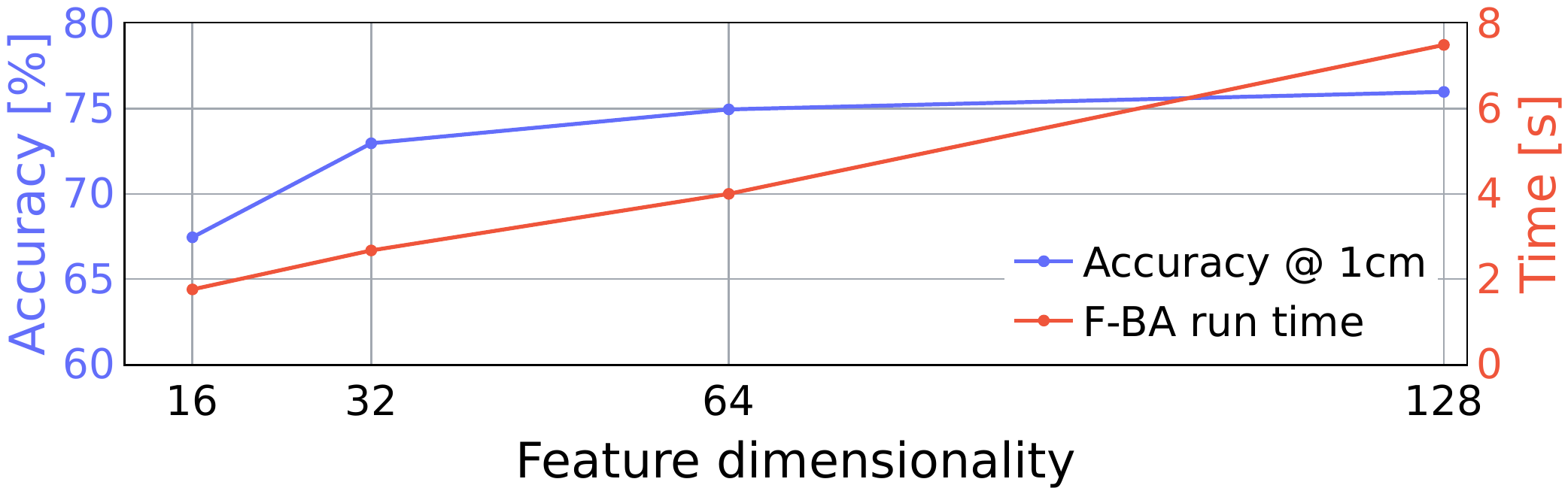}
    \caption{\textbf{Impact of the feature dimensionality.} %
    Dense features computed by S2DNet can be naively reduced to accelerate the featuremetric bundle adjustment by 2 while incurring only a minor drop of triangulation accuracy.
    }%
    \label{fig:feature-dimension}%
\end{figure}

\subsection{Image resolution}
The image resolution at which the dense features are extracted has a large impact on the accuracy of the refinement.
In Figure~\ref{fig:image-resolution} we quantify in the impact on both triangulation accuracy and run time for the ETH3D \emph{Courtyard} scene (38 images).
The accuracy drops significantly when the resolution is smaller than 1600$\times$1066px, which amounts to 25\% of the full image resolution.
Doubling the resolution to 3200$\times$2132px yields noticeable improvements, albeit significantly increases the extraction time and the consumption of GPU VRAM.
As a reference, extracting only fine-level S2DNet features (4 convolutions) from 3200$\times$2132px images requires around 10GB of GPU VRAM.

\subsection{Reference selection for keypoint adjustment}
Selecting some observations as references is necessary to avoid the drift.
In a given track, the keypoint adjustment selects the point that is the most connected (topological center), while the bundle adjustment selects the point closest to the robust mean in feature space (feature center).
Could we use the feature center for selecting the reference of the keypoint adjustment?
By minimizing the feature distance to this unique reference, we could reduce the number of residuals from quadratic (pairwise constraints) to linear (unary constraints) and thus accelerate the optimization.

Retaining pairwise constraints however allows the optimization to separate tracks that were incorrectly merged by the track separation algorithm. 
This is not necessary in the bundle adjustment, as tracks are already filtered by the robust geometric estimation and can thus be assumed to be correct, but is common for unverified track.
We evaluate the impact of the reference selection in the keypoint adjustment and report the results in Table~\ref{tab:eth3d-ablation_supp}. 
For both SuperPoint and D2-Net, using the feature center results in lower completeness and accuracy than the topological center.
It also results in a lower track length, which confirms that the topological reference allows to retain incorrectly-merged tracks.
Since the feature center still performs relatively well, it could be considered in case of tighter computational constraints.

Furthermore, Table~\ref{tab:eth3d-ablation_supp} highlights the importance of the featuremetric keypoint adjustment. 
The benefits are larger for D2-Net, which detects very noisy keypoints.
As a consequence, many correct albeit noisy matches are rejected by the geometric verification.
Our keypoint adjustment not only allows more points to be triangulated, thus increasing the completeness of the model, but also increases the accuracy of the triangulated points.

\begin{table}[t]
\centering
\footnotesize{\setlength\tabcolsep{3.2pt}\
\begin{tabular}{llcccccccc}
    \toprule
    &\multirow{2}{1.7cm}[-.3em]{Triangulation $\drsh$~Refinement}
    & \multicolumn{2}{c}{Acc.\ (\%)} & \multicolumn{3}{c}{Compl.\ (\%)} & \multirow{2}{0.8cm}[-.3em]{track length}\\
    \cmidrule(lr){3-4}
    \cmidrule(lr){5-7}
    && 1cm & 2cm & 1cm & 2cm & 5cm\\
    \midrule
    \multirow{5}{*}{\begin{sideways} SuperPoint \end{sideways}}
    & unrefined & 18.03 & 31.97 & 0.07 & 0.49 & 5.03 & 4.17\\
    &$\drsh$ Patch Flow~\cite{dusmanu2020} & 37.00 & 55.18 & 0.15 & 0.93 & 7.44 & \b{5.24}\\
    &$\drsh$ F-BA & 43.65 & 62.44 & 0.18 & 1.06 & 7.70 & 4.17 \\
    &$\drsh$ +F-KA (feat-ref) & 45.05 & 64.84 & 0.18 & 1.12 & 7.76 & 4.88 \\
    &$\drsh$ +F-KA (topol-ref) & \b{46.46} & \b{65.41} & \b{0.19} & \b{1.14} & \b{8.19} & 5.02\\
    \midrule
    \multirow{5}{*}{\begin{sideways} D2-Net \end{sideways}}
    & unrefined & 7.68 & 13.98 & 0.02 & 0.17 & 2.19 & 3.29 \\
    &$\drsh$ Patch Flow~\cite{dusmanu2020} & 34.64 & 52.36 & 0.16 & 1.00 & 8.10 & \b{4.99} \\
    &$\drsh$ F-BA & 39.30 & 58.59 & 0.15 & 0.94 & 6.99 & 3.29 \\
    &$\drsh$ +F-KA (feat-ref) & 43.35 & 62.54 & 0.19 & 1.18 & 8.36 & 4.49 \\
    &$\drsh$ +F-KA (topol-ref) & \b{44.21} & \b{64.22} & \b{0.20} & \b{1.20} & \b{8.72} & 4.63\\
    \bottomrule
\end{tabular}}
\vspace{.04in}
\caption{\textbf{Additional ablation study on ETH3D Facade.}
i)~Featuremetric keypoint adjustment significantly improves the completeness, especially for noisy keypoints as in D2-Net.
ii)~Keypoint adjustment against the topological center in each tentative track (topol-ref) improves the point cloud in accuracy and completeness over KA towards the robust feature center (feat-ref) because it allows to merge tracks.
}
\label{tab:eth3d-ablation_supp}
\end{table}

\subsection{Number of feature levels}
Using multiple feature levels enlarges the basin of convergence but increases the computational requirements.
The radius of convergence that is required depends on the noise of the keypoint detector and on the resolution of the image from which keypoints are detected.
When performing detection and refinement at identical image resolutions, the optimal displacement is at most a few pixels for most keypoint detectors.
In this case, the fine level of S2DNet feature maps is sufficient.
We empirically measured that its radius of convergence is approximately 3 pixels, although the multiview constraints enable to refine over much larger distances.

We thus use a single feature level for all experiments involving SIFT, SuperPoint, and R2D2.
D2-Net require a different treatment, as its detection noise is significantly larger.
This is partly due to the aggressive downsampling of its CNN backbone and to the low resolution of its output heatmap.
As a consequence, we employ both fine and medium feature levels for D2-Net.
Both keypoint and bundle adjustments run the optimization successively at the coarser and finer levels.

\subsection{Dimensionality of the features}
Throughout this paper, we used 128-dimensional dense features extracted by S2DNet~\cite{germain2020s2dnet}.
Relying on compact features would easily reduce the memory footprint and the run time of the refinement.
To demonstrate these benefits, we show in Figure~\ref{fig:feature-dimension} the relationship between the dimension, the run time of the BA, and the triangulation accuracy when retaining only the first $k$ channels of the S2DNet features.
Features with fewer dimensions yield a faster refinement.
The accuracy drops moderately but we expect a smaller reduction with features explicitly trained for smaller dimensions.

\begin{figure*}[t]
    \centering
    \begin{minipage}{0.015\textwidth}
    \rotatebox[origin=c]{90}{SuperPoint}
    \end{minipage}%
    \hspace{1mm}%
    \begin{minipage}{0.95\textwidth}
        \centering
        \includegraphics[width=\linewidth]{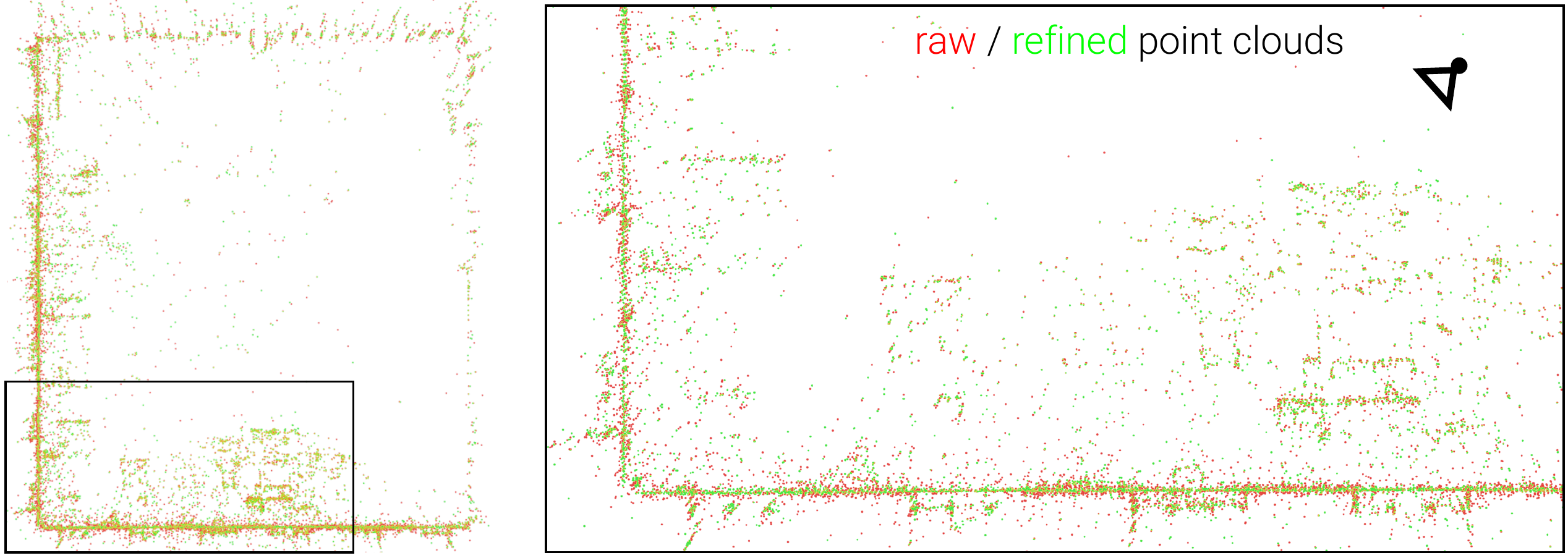}
    \end{minipage}
    \vspace{1mm}
    
    \begin{minipage}{0.015\textwidth}
    \rotatebox[origin=c]{90}{Accuracy - unrefined}
    \end{minipage}
    \hspace{1mm}%
    \begin{minipage}{0.40\textwidth}
    \includegraphics[width=\linewidth]{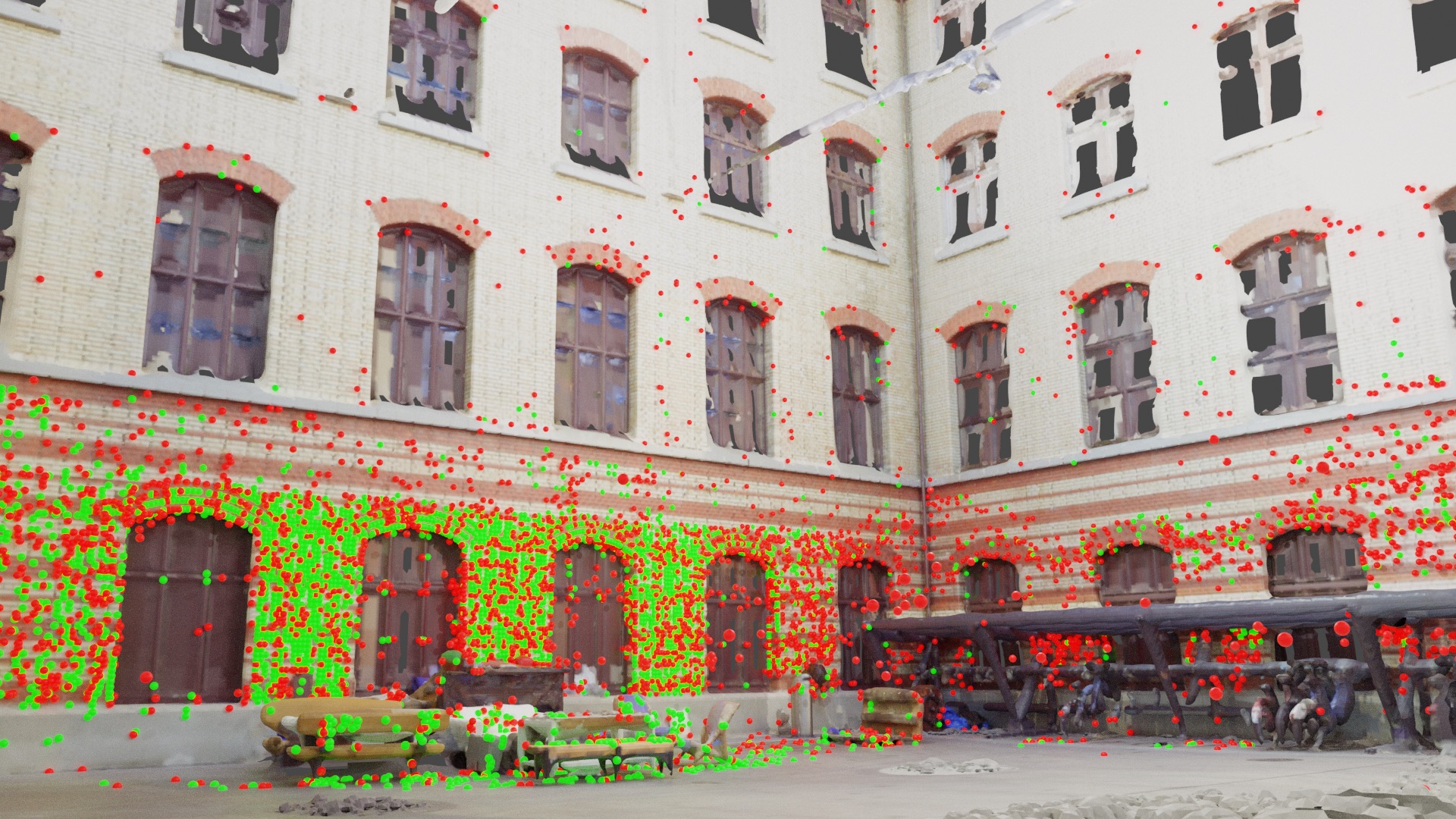}
    \end{minipage}
    \hspace{5mm}%
    \begin{minipage}{0.015\textwidth}
    \rotatebox[origin=c]{90}{Accuracy - refined}
    \end{minipage}
    \hspace{1mm}%
    \begin{minipage}{0.40\textwidth}
    \includegraphics[width=\linewidth]{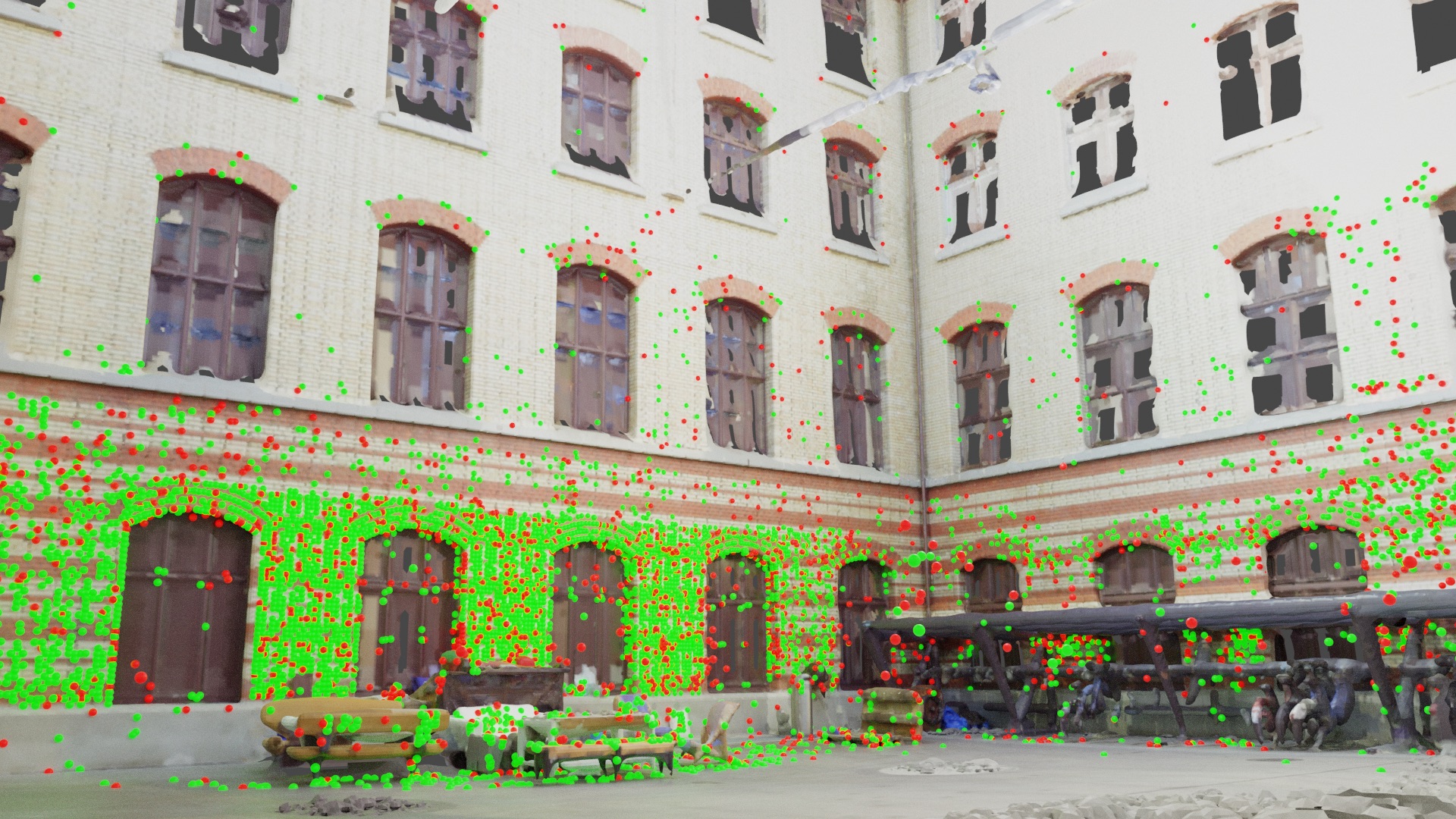}
    \end{minipage}
    \vspace{5mm}
    
    \begin{minipage}{0.015\textwidth}
    \rotatebox[origin=c]{90}{D2-Net}
    \end{minipage}%
    \hspace{1mm}%
    \begin{minipage}{0.95\textwidth}
        \centering
        \includegraphics[width=\linewidth]{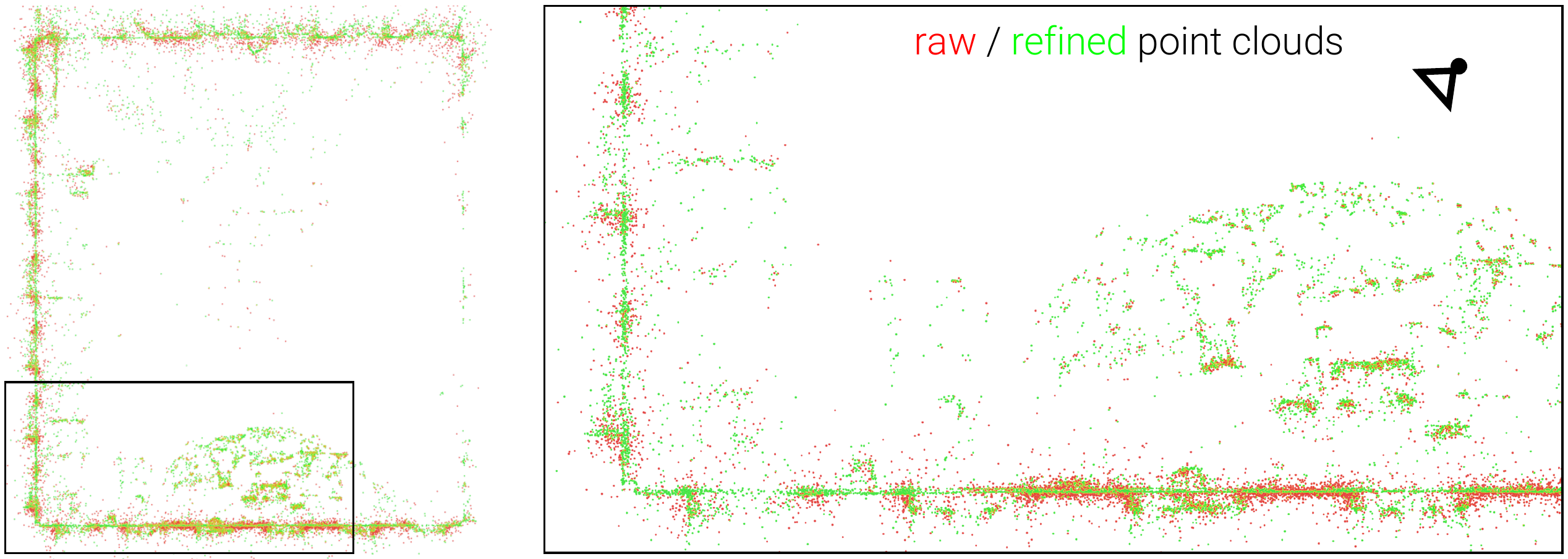}
    \end{minipage}%
    \vspace{1mm}
    
    \begin{minipage}{0.015\textwidth}
    \rotatebox[origin=c]{90}{Accuracy - unrefined}
    \end{minipage}
    \hspace{1mm}%
    \begin{minipage}{0.40\textwidth}
    \includegraphics[width=\linewidth]{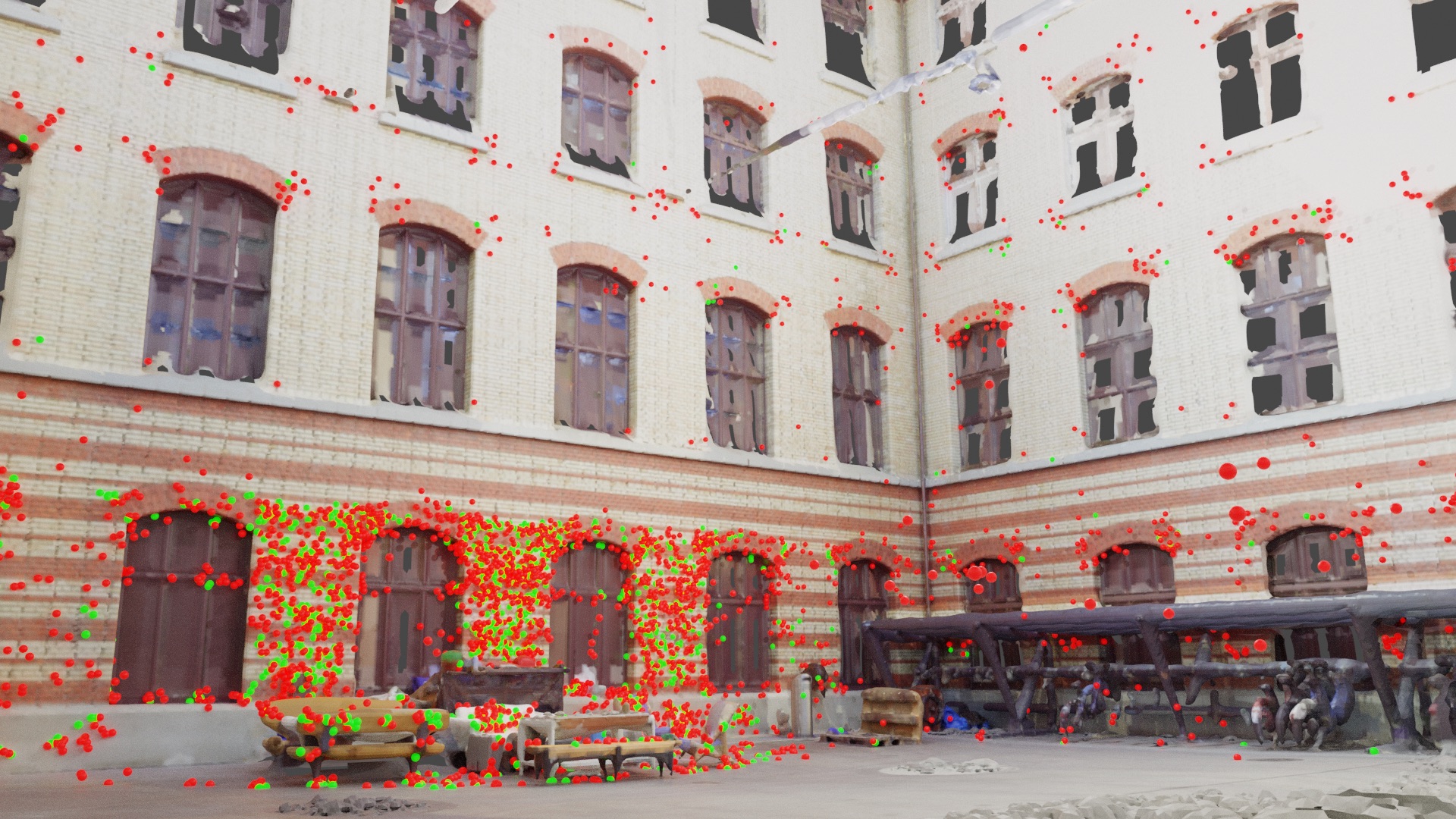}
    \end{minipage}
    \hspace{5mm}%
    \begin{minipage}{0.015\textwidth}
    \rotatebox[origin=c]{90}{Accuracy - refined}
    \end{minipage}
    \hspace{1mm}%
    \begin{minipage}{0.40\textwidth}
    \includegraphics[width=\linewidth]{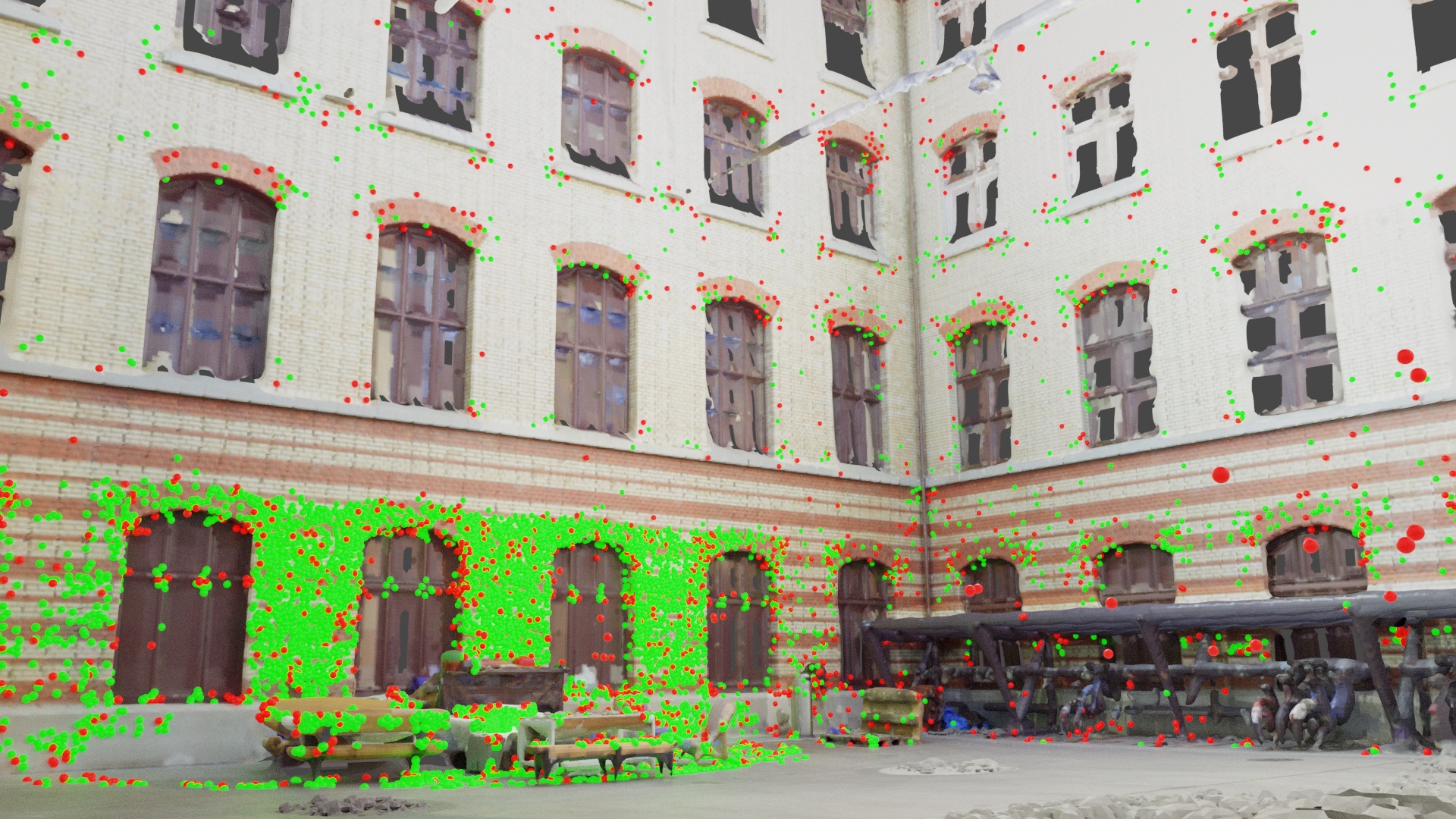}
    \end{minipage}
    \vspace{-1mm}
    
    \caption{\textbf{Refinement on ETH3D \emph{Courtyard}.}
    In the top parts, we show for both SuperPoint (top) and D2-Net (bottom) top-down views of the sparse point clouds triangulated with raw (in \red{red}) and refined (in \green{green}) keypoints.
    The refined point clouds better fit the geometry of the scene, especially on planar walls.
    In the lower parts, we also show images in which points are colored as accurate (in \green{green}) or inaccurate (in \red{red}) at 1cm for raw (left) and refined (right) point clouds.
    }
    \label{fig:triangulation:courtyard}%
\end{figure*}

\section{Cost map approximation}
\label{section:supp:costmap}
We mention in Section~\ref{section:implementation} that the memory efficiency of the bundle adjustment can be improved by precomputing the featuremetric cost.
We provide here more details.

\PAR{Description:}
Given $D$-dimension features, the featuremetric bundle adjustment (Eq.~\ref{eq:fba}) involves residuals and Jacobian matrices of dimension $D$.
Unlike the keypoint adjustment, which can optimize tracks independently, all bundle parameters are updated simultaneously and the memory requirements are thus prohibitive.
Given the 2D reprojection $\*p_{ij} = \Pi\left(\*R_i\*P_j+\*t_i, \*C_i\right)$, this formulation loads in memory the dense features $\*F_i$, interpolates them at $\*p_{ij}$, and compute the residuals $\*r_{ij} = \*F_i\left[\*p_{ij}\right] - \*f^j $ for the cost $E_{ij} = \left\lVert\*r_{ij}\right\rVert_\gamma$.

To reduce the memory footprint, we can exhaustively precompute patches of feature distances and treat them as one-dimensional residuals $\bar{\*r}_{ij} = \left\lVert\*F_i - \*f^j\right\rVert\left[\*p_{ij}\right]$.
The cost then becomes $\bar{E}_{ij} = \gamma(\bar{\*r}_{ij})$.
Such distances only need to be computed once since the reference $\*f^j$ is kept fixed throughout the optimization.
This precomputed cost reduces the peak memory by a factor $D$, with often $D{=}128$.
It is similar to the Neural Reprojection Error recently introduced by Germain~\etal~\cite{Germain_2021_CVPR} for camera localization.

\PAR{Analysis:}
Swapping the distance computation and the sparse interpolation introduces an approximation error.
We first write the bilinear or bicubic interpolation as a sum over features $\*F_k$ on the discrete grid:
\begin{equation}
    \*F\left[\*p\right] = \sum_k w_k\*F_k \quad \text{with} \quad \sum_k w_k = 1 \enspace.
\end{equation}
We assume that the features are L2-normalized $\left\lVert\*F_k\right\rVert = 1$, such that $\left\lVert\*F\left[\*p\right]\right\rVert \approx 1$.
For a squared loss function, the approximation error can then be written as:
\begin{multline}
    \left\lVert\*F - \*f\right\rVert^2[\*p] - \left\lVert\*F[\*p]-\*f\right\rVert^2\\
    \approx 1 - \left\lVert\*F[\*p]\right\rVert^2
    = \frac{1}{2}\sum_k\sum_lw_kw_l \left\lVert\*F_k - \*F_l\right\rVert^2 \enspace.
\end{multline}
This error is zero at points on the discrete grid and increases with the roughness of the feature space.
This approximation thus displaces the local minimum of the cost by at most 1 pixel but most often by much less.

\PAR{Improvement:}
This approximation however degrades the correctness of the approximate Hessian matrix that the Levenberg-Marquardt algorithm~\cite{levenberg1944method} relies on for fast convergence.
We found that also optimizing the squared spatial derivatives of this cost significantly improves the convergence.
This simply amounts to augmenting the scalar residual map with dense derivative maps:
\begin{equation}
    \tilde{\*r}_{ij} = 
    \begin{pmatrix}
        \left\lVert\*F_i - \*f^j\right\rVert \\
        \frac{\partial\left\lVert\*F_i - \*f^j\right\rVert}{\partial x}\\
        \frac{\partial\left\lVert\*F_i - \*f^j\right\rVert}{\partial y}
    \end{pmatrix}\left[\*p_{ij}\right] \enspace.
\end{equation}
This improvement results in three-dimensional residuals, which is still smaller than $D$ when $D{=}128$.
Using the spatial derivatives, we can also compute an exact, more accurate bicubic spline interpolation of the cost landscape.

\begin{table}[t]
\centering
\footnotesize{\setlength\tabcolsep{3.2pt}\
\begin{tabular}{lcccccccc}
    \toprule
    \multirow{2}{1.7cm}[-.3em]{SuperPoint $\drsh$~Refinement}
    & \multicolumn{2}{c}{Acc.\ (\%)} & \multicolumn{2}{c}{Compl.\ (\%)} & \multirow{2}{0.9cm}[-.3em]{\centering Time \\ (s)} & \multirow{2}{1.0cm}[-.3em]{\centering Memory (GB)}\\
    \cmidrule(lr){2-3}
    \cmidrule(lr){4-5}
    & 1cm & 2cm & 1cm & 2cm\\
    \midrule
    unrefined & 64.27 & 76.47 & 0.37 & 1.44 & - & -\\
    $\drsh$ ours (exact) & \b{81.31} & \b{88.50} & \b{0.47} & \b{1.74} & 42.22 & 7.3 \\
    $\drsh$ ours (cost maps) & 80.27 & 87.81 & \b{0.47} & 1.72 & \b{29.86} & \b{0.15}\\
    \bottomrule
\end{tabular}}
\vspace{.04in}
\caption{\textbf{Triangulation with cost map approximations.}
Using precomputed cost maps increase the efficiency of the bundle adjustment with a marginal loss of accuracy.
}
\label{tab:eth3d-costmaps}
\end{table}

\PAR{Evaluation:}
We now show experimentally that this approximation often does not, or only minimally, impairs the accuracy of the refinement.
Table~\ref{tab:eth3d-costmaps} reports the results of the triangulation of SuperPoint features on the ETH3D dataset.
The approximation reduces the accuracy by less than 1\% and does not alter the completeness.
It however significantly reduces the memory consumption of the bundle adjustment, allowing it to scale to thousands of images.
Note that all experiments in Sections~\ref{section:triangulation}, \ref{section:pose}, and \ref{section:phototourism} do not use the cost map approximation as the corresponding scenes are relatively small.

\section{Experimental details}
\label{section:supp:details}
\subsection{ETH3D - Sections~\ref{section:triangulation} and \ref{section:pose}}
For the experiments on ETH3D, we use the evaluation code provided by Dusmanu \etal~\cite{dusmanu2020}.
We use the original implementations of SuperPoint~\cite{superpoint}, D2-Net~\cite{dusmanu2019d2}, and R2D2~\cite{revaud2019r2d2}, and extract root-normalized SIFT~\cite{lowe2004distinctive} features using COLMAP~\cite{schoenberger2016sfm}.
For both sparse and dense feature extraction, the images are resized so that their longest dimension is equal to 1600 pixels.
The tentative matches are filtered according to the recipe described in~\cite{dusmanu2020}.

\subsection{Structure-from-Motion - Section~\ref{section:phototourism}}
We tune the hyperparameters on the training scenes \emph{Temple Nara Japan}, \emph{Trevi Fountain}, and \emph{Brandenburg Gate}.
The results in the main paper are computed on the test scenes \emph{Sacre Coeur}, \emph{Saint Peter's Square}, and \emph{Reichstag}, using the data and code provided by the challenge organizers.

For SIFT~\cite{lowe2004distinctive}, we use the mutual check, a ratio test with threshold 0.85 for the multi-view and 0.9 for the stereo tasks, and DEGENSAC with an inlier threshold of 0.5px.
For D2-Net~\cite{dusmanu2019d2}, we use the mutual check and inlier thresholds of 2px and 0.5px for raw and refined keypoints, respectively.
For SuperPoint+SuperGlue~\cite{superpoint, sarlin2020superglue}, we do not use additional match filtering and we select an inlier thresholds of 1.1px and 0.5px for raw and refined keypoints, respectively.
All sparse local and dense features are extracted at full image resolution, which is generally not larger than 1024px.

\subsection{Ablation study - Section~\ref{section:ablation}}
The triangulation metrics are reported for the ETH3D scene \emph{Facade}, which is the largest with 76 images.
We use SuperPoint local features as they perform best in all earlier experiments and we store dense feature maps in every experiment.
The localization AUC is measured over all 13 scenes in ETH3D with 10 holdout images per scene. We now detail the different baselines.

Localization is achieved in ``F-KA'' by first refining the keypoints, triangulating the map and finally performing query keypoint adjustment as described in section \ref{subection:supp:camerapose}. For localization with ``F-BA'', we refined the triangulated model using featuremetric bundle adjustment and then refined the pose from PnP+RANSAC using qBA.

In the entry ``w/ F-BA drift'', we use the robust reference (Eq.~\ref{eq:reference}) to select the observation in each track which is most similar to the robust reference as the source frame. 
The optimizer then minimizes the error between each other observation and the current, moving reference of the source frame.
Since only the index of the source frame is fixed during the optimization, this method does not account for drift, which appears to yield higher accuracy but suffers from repeatability problems during localization.

The baseline ``PatchFlow + F-BA'' uses the keypoint refinement from Dusmanu \etal~\cite{dusmanu2020} as initialization, and runs our featuremetric bundle adjustment on top of it. We used the exact same parameters for PatchFlow as presented in~\cite{dusmanu2020}.

The entry ``higher resolution'' corresponds to input images at double the resolution than all the other experiments, i.e. 3200 pixels in the longest dimension.

For the ``photometric'' baseline, we use RGB images (while Woodford \etal~\cite{woodford2020large} use grayscale images), we warp patches of 4$\times$4 pixels at the featuremap resolution (1600 pixels in the longest dimension) with fronto-parallel assumption, and apply normalized cross correlation (NCC).
Identically to our featuremetric BA and to LSPBA~\cite{woodford2020large}, the source frame is selected as the observation closest to the robust mean.

We report results for dense features extracted from a VGG-16 CNN, trained on ImageNet~\cite{deng2009imagenet}, at the layer \texttt{conv1\_2} (64 channels) and for the fine feature map predicted by PixLoc~\cite{sarlin21pixloc} (32 channels).
The model of PixLoc, trained on MegaDepth~\cite{li2018megadepth}, was kindly provided by its authors.
In DSIFT~\cite{liu2010sift} (128 channels), we apply a bin size of 4 and a step size of 1 and refer to the VLFeat implementation~\cite{vedaldi10vlfeat} for more details.

\subsection{Scalability}
All experiments were conducted on 8 CPU cores (Intel Xeon E5-2630v4) and one NVIDIA RTX 1080 Ti. 
The subsets from the Aachen Day-Night v1.1 model~\cite{sattler2018benchmarking,sattler2012image,Zhang2020IJCV} were selected as the images with the largest visibility overlap, in descending order. 
To accelerate the feature matching, each image was matched only to its top 20 most covisible reference images in the original Aachen SfM model.
We use SuperPoint~\cite{superpoint} features and match image pairs with the mutual check and distance thresholding at 0.7. 
During BA, we apply the sparse Schur solver from Ceres for each linear system in LM, while we use sparse Cholesky in KA, similar to~\cite{dusmanu2020}. 
Featuremetric bundle adjustment is stopped after 30 iterations while KA runs for at most 100 iterations and stops when parameters change by less than $10^{-4}$.

To refine the full Aachen Day-Night model, we use SuperPoint features matched with SuperGlue~\cite{sarlin2020superglue} from the Hierarchical Localization toolbox~\cite{hloc, sarlin2019coarse}. We refine the keypoints with KA, then triangulate the points with fixed poses from the reference model.
Finally, we run a full bundle adjustment of the model with the proposed approximation by cost maps.
    \fi
\else
    
\fi

\clearpage
{\small
\bibliographystyle{ieee_fullname}
\bibliography{mybib}
}

\end{document}